\documentclass[11pt]{article}

\usepackage[final]{acl}

\usepackage{times}
\usepackage{latexsym}

\usepackage[T1]{fontenc}

\usepackage[utf8]{inputenc}

\usepackage{microtype}

\usepackage{inconsolata}

\usepackage{graphicx}

\usepackage{amsfonts}
\usepackage{amsmath}
\usepackage{booktabs}
\usepackage{pifont}
\usepackage{wrapfig}
\usepackage{multirow}
\usepackage{graphicx}
\usepackage{wrapfig}
\usepackage[table]{xcolor}
\usepackage{tcolorbox}
\tcbuselibrary{breakable}
\usepackage{fontawesome}
\definecolor{aliceblue}{rgb}{0.88, 0.95, 1.0}
\definecolor{grey}{rgb}{0.5, 0.5, 0.5}

\usepackage{listings}
\usepackage{xcolor} 

\lstdefinestyle{jsonstyle}{
    basicstyle=\ttfamily\small,
    stringstyle=\color{purple},
    keywordstyle=\color{blue},
    showstringspaces=false,
    breaklines=true,         
    frame=none,            
    literate=
      *{0}{{{\color{red}0}}}{1}
      {1}{{{\color{red}1}}}{1}
      {2}{{{\color{red}2}}}{1}
      {3}{{{\color{red}3}}}{1}
      {4}{{{\color{red}4}}}{1}
      {5}{{{\color{red}5}}}{1}
      {6}{{{\color{red}6}}}{1}
      {7}{{{\color{red}7}}}{1}
      {8}{{{\color{red}8}}}{1}
      {9}{{{\color{red}9}}}{1},
}

\lstdefinestyle{xmlstyle}{
    basicstyle=\ttfamily\small,
    showstringspaces=false,
    breaklines=true,            
    frame=none,               
    morestring=[s]{<}{>},
    stringstyle=\color{blue},
}

\title{\raisebox{-1.2mm}{\includegraphics[width=0.0525\textwidth]{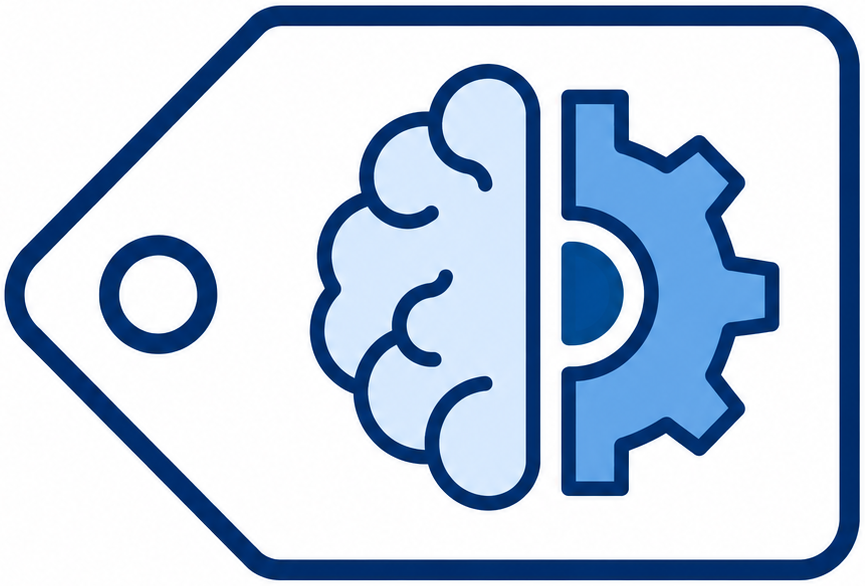}} TagPR: Tag-Guided Process Supervision for Personalization Reasoning in Large Language Models}

\author{
  \textbf{Song Jin}\textsuperscript{1}, \textbf{Juntian Zhang}\textsuperscript{1}, \textbf{Ruyu Lyu}\textsuperscript{1}, \textbf{Yong Liu}\textsuperscript{1}, \\ \textbf{Xun Zhang}\textsuperscript{2}, \textbf{Yufei Zhang}\textsuperscript{2}, \textbf{Fei Jiang}\textsuperscript{2}, \textbf{Guojun Yin}\textsuperscript{2}, \textbf{Wei Lin}\textsuperscript{2}, \textbf{Rui Yan}\textsuperscript{3}\thanks{\ \ Corresponding author.} \\
  \textsuperscript{1}Renmin University of China, \textsuperscript{2}Meituan, \textsuperscript{3}Wuhan University\\
  jinsong8@ruc.edu.cn
}

\begin{document}
\maketitle
\begin{abstract}
Recent advancements have endowed Large Language Models with impressive general reasoning capabilities. However, these reasoning models often perform worse than non-reasoning models on personalization tasks. While some methods use outcome-based RL to improve personalization reasoning, they fail to supervise the reasoning process. As a result, models may reach correct answers through flawed reasoning chains, limiting further improvement. To address this, we propose TagPR, a novel framework that adds semantic tags to the reasoning process for step-by-step guidance. TagPR first automatically generates a structured, tagged dataset for Supervised Fine-Tuning. It then employs a multi-stage RL process guided by a composite reward signal, which integrates tag-based process supervision with a novel Personalization Reward Model with User Embeddings to achieve fine-grained alignment with user-specific logic. Extensive experiments on public LaMP, LongLaMP, PGraphRAG, and a self-constructed dataset demonstrate that our approach achieves state-of-the-art results, delivering an average improvement of 32.65\% over the base model across all LaMP benchmark tasks. Our work demonstrates that tag-guided process supervision is an effective approach for personalization reasoning.
\end{abstract}

\section{Introduction}

Research on large-model reasoning now spans mathematics, coding, and multimodal domains~\citep{guo2025deepseek, yu2026dapo, jin2026diningbench, zhang2026viper}, yet these advances do not readily translate to personalization tasks. This domain is increasingly critical, powering a broad range of real-world applications, including personalized recommendation, intelligent writing assistance, targeted advertising, and user-tailored conversational agents. Effective personalization demands far more than generic logic. It requires personalization reasoning: the ability to carefully analyze a user's historical data, infer their unique preferences and thought patterns, and synthesize this understanding to generate responses that are precisely tailored to each user.

Surprisingly, even state-of-the-art reasoning-centric LLMs struggle in this domain, frequently failing to surpass non-reasoning baselines on personalization benchmarks. This performance deficit stems from a fundamental misalignment: models optimized for general-purpose reasoning inherently prioritize their internal, generalized logic over the
user's profile. 
Consequently, they generate responses that are either overly generic or directly contradict established user preferences.
Recent pioneering works, such as R2P~\citep{luo2025reasoning} and RPM~\citep{kim2026rpm}, have identified this limitation. Although these approaches mitigate the issue by employing prompt templates or pre-constructed reasoning paths, they function merely as external scaffolds rather than fundamentally improving the model's intrinsic capacity for user-centric reasoning.

\begin{figure*}
 \centering
\includegraphics[width=0.97\linewidth]{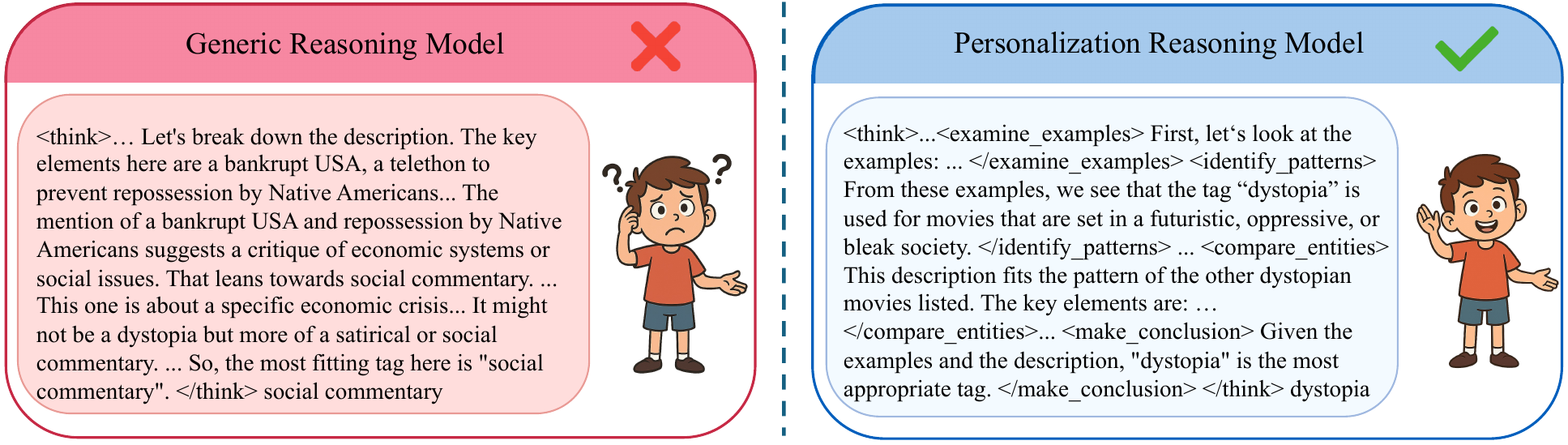}
\caption{A comparison of reasoning paths. Left: The Generic Reasoning Model (Qwen3-8B) uses free-form logic, leading to an incorrect tag (``social commentary''). Right: Our Personalization Model follows a structured path to correctly infer the user-specific tag (``dystopia'').}
\label{fig:reasoning_comparison}
\end{figure*}

Our core motivation stems from the observation that personalization reasoning is not a monolithic act of intuition but a structured process with multiple steps involving the analysis of user history, the identification of recurring patterns, and the application of those patterns to new contexts. The opaque and unconstrained reasoning of standard LLMs is poorly suited for this task.
Furthermore, existing RL-based approaches for eliciting personalization reasoning are often hampered by their reliance on outcome-based rewards~\citep{zhang2025prlm, li2025extended}. Such signals are inherently sparse because a correct final answer does not necessarily imply a valid reasoning process, which 
limits further improvement. The ability to provide structured guidance signals within the model's internal reasoning logic therefore becomes highly valuable. To this end, we introduce TagPR.
Instead of allowing the model to reason implicitly, we compel it to externalize its logic into a sequence of discrete, interpretable steps, each marked with a semantic tag (e.g., \texttt{<examine\_examples>}, \texttt{<identify\_patterns>}). These tags act as cognitive waypoints, transforming the complex task of personalization into a manageable, explicit procedure that the model can learn to execute robustly, as illustrated in Figure~\ref{fig:reasoning_comparison}.
By transforming implicit reasoning steps into an explicit structure, our approach enables precise process-level supervision based on tags embedded in the reasoning trajectory during the RL stage.

This is achieved through a synergistic training strategy. First, we pioneer a data-driven pipeline to automatically generate a new dataset of reasoning chains labeled with these semantic tags. We use this dataset for Supervised Fine-Tuning (SFT) to instill the foundational grammar of structured and personalized thought. Following this, we employ a multi-stage RL process to refine this capability. This RL phase is guided by a novel composite reward that combines tag-based structural constraints with a fine-grained signal from our new Personalization Reward Model with User Embeddings (PRMU), which explicitly aligns the model's reasoning with user-specific logic. Our key contributions are threefold:

\MakeUppercase{\romannumeral 1}. We formulate personalization reasoning as a sequence of semantically tagged actions and introduce a data-driven pipeline to automatically generate and label reasoning chains with semantic tags.
    
\MakeUppercase{\romannumeral 2}. We introduce a synergistic SFT and multi-stage RL training framework. This process is guided by a unique composite reward signal that integrates tag-based constraints and our novel Personalization Reward Model with User Embeddings (PRMU) for fine-grained alignment with user logic.
    
\MakeUppercase{\romannumeral 3}. We demonstrate through extensive experiments on public LaMP, LongLaMP, PGraphRAG and a self-constructed dataset that 
\textbf{TagPR} achieves state-of-the-art results.

\section{Related Work}

\begin{figure*}
 \centering
 \includegraphics[width=0.97\linewidth]{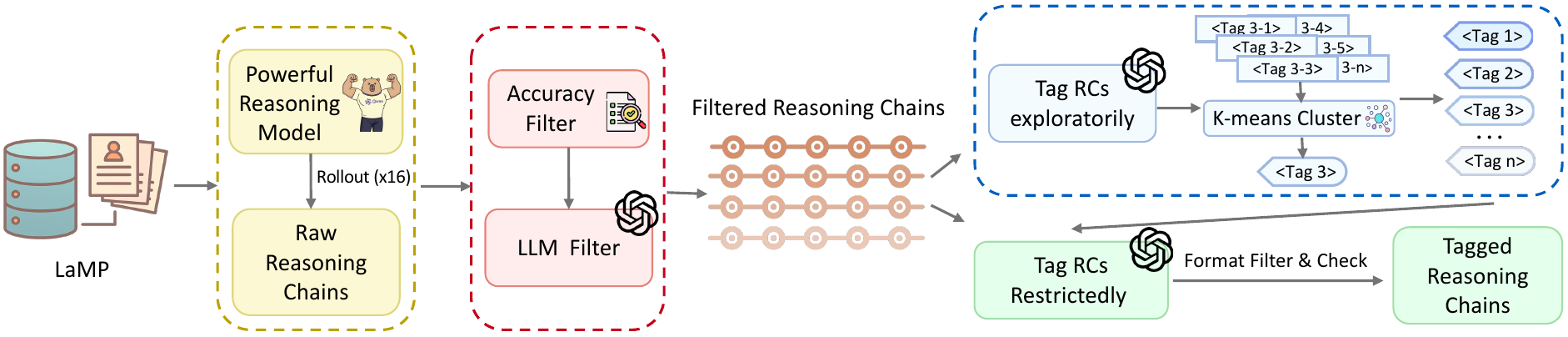}
 \caption{The pipeline for constructing our Tagged Reasoning Chains dataset. The process includes raw chains generation from LaMP, a two-stage quality filter, and a two-phase tagging procedure where primary tags are first defined via clustering and then applied in a restricted final annotation.}
 \label{fig:data}
\end{figure*}

\textbf{Large Language Model Personalization} 
LLM personalization has evolved rapidly.
A dominant approach is retrieval-augmented generation, with innovations including feedback-optimized retrieval~\citep{salemi2024optimization} and generation-calibrated retrievers~\citep{mysore2024pearl}. PAG~\citep{richardson2023integrating} enhances retrieval by integrating user history summarization. Beyond retrieval, research has explored core personalization components~\citep{wu2024understanding}.
DPL~\citep{qiu2025measuring} models inter-user differences to capture unique preferences. Parameter-efficient approaches include OPPU~\citep{tan2024democratizing} with user-specific lightweight modules, PER-PCS~\citep{tan2024personalized} for collaborative PEFT sharing,
plug-and-play user embeddings (PPlug)~\citep{liu2025llms+}, and HYDRA~\citep{zhuang2024hydra} for black-box personalization. Additional methods include multi-stage decomposition~\citep{li2023teach} and multi-objective parameter merging (P-Soups)~\citep{jang2023personalized}.
Recent work has further expanded personalization research toward personalized user-agent simulation~\citep{jin2025beyond} and user-specific evaluation~\citep{qiu2026preference}.

\noindent \textbf{Personalization Reasoning} 
Personalization reasoning represents an emerging intersection of reasoning capabilities and personalization tasks. Early approaches primarily use prompting strategies for black-box models: RPM~\citep{kim2026rpm} constructs individualized reasoning paths from user history, while R2P~\citep{luo2025reasoning} employs hierarchical reasoning templates. Fine-tuning approaches include generating reasoning paths followed by iterative self-training~\citep{salemi2025reasoning}, and reinforcement learning for preference inference through extended inductive reasoning~\citep{li2025extended}. Most closely related to our work, PrLM~\citep{zhang2025prlm} uses contrastive reward models with reinforcement learning for reasoning in personalization generation tasks. While these methods have made notable progress, they typically address personalization reasoning through either template-guided generation or reward-based optimization without fundamentally restructuring how models approach the multi-faceted nature of personalization tasks.

\section{Methodology}
This section presents the methodology for \textbf{TagPR}. We formulate the task and overview the framework in Section~\ref{sec:task_formulation}, detail the reasoning structure initialization in Section~\ref{sec:structure_init}, and present the Tag-Guided Reinforcement Learning in Section~\ref{sec:tag_guided_rl}.

\subsection{Task Formulation \& Overview}
\label{sec:task_formulation}
We define personalized reasoning as the task of generating a user-specific response $y$ to a query $x$, conditioned on the user's profile $P_u = \{(x_i, y_i)\}_{i=1}^{N_u}$, where $N_u$ is the total number of historical records available for user $u$, and each record $(x_i, y_i)$ is a past input–output pair.
To incorporate reasoning into this process, the task is extended to first generate an explicit reasoning chain $c$ before producing the final response $y$. Conditioned on the query $x$ and a relevant profile subset $p_u \subseteq P_u$, the objective for a model parameterized by $\theta$ is to maximize the joint likelihood of the reasoning chain and the response:
\begin{equation}
\label{eq:joint_likelihood}
\resizebox{0.89\linewidth}{!}{$
p(c, y | x, p_u; \theta) = p(c | x, p_u; \theta) \cdot p(y | c, x, p_u; \theta).
$}
\end{equation}
The core challenge is to ensure that the reasoning chain $c$ is coherent and faithful to the user's profile $p_u$, while the response $y$ remains consistent with this explicit reasoning.

The overall \textbf{TagPR} framework addresses this challenge through two sequential phases. First, a lightweight \textit{Reasoning Structure Initialization} stage injects the syntactic grammar of tagged reasoning into the model via a small amount of high-quality supervised data. Second, a \textit{Tag-Guided Reinforcement Learning} stage refines the model's personalization reasoning through tag-based process supervision and fine-grained user-aware reward signals. 

\begin{figure*}[t]
    \centering
    \includegraphics[width=1.0\linewidth]{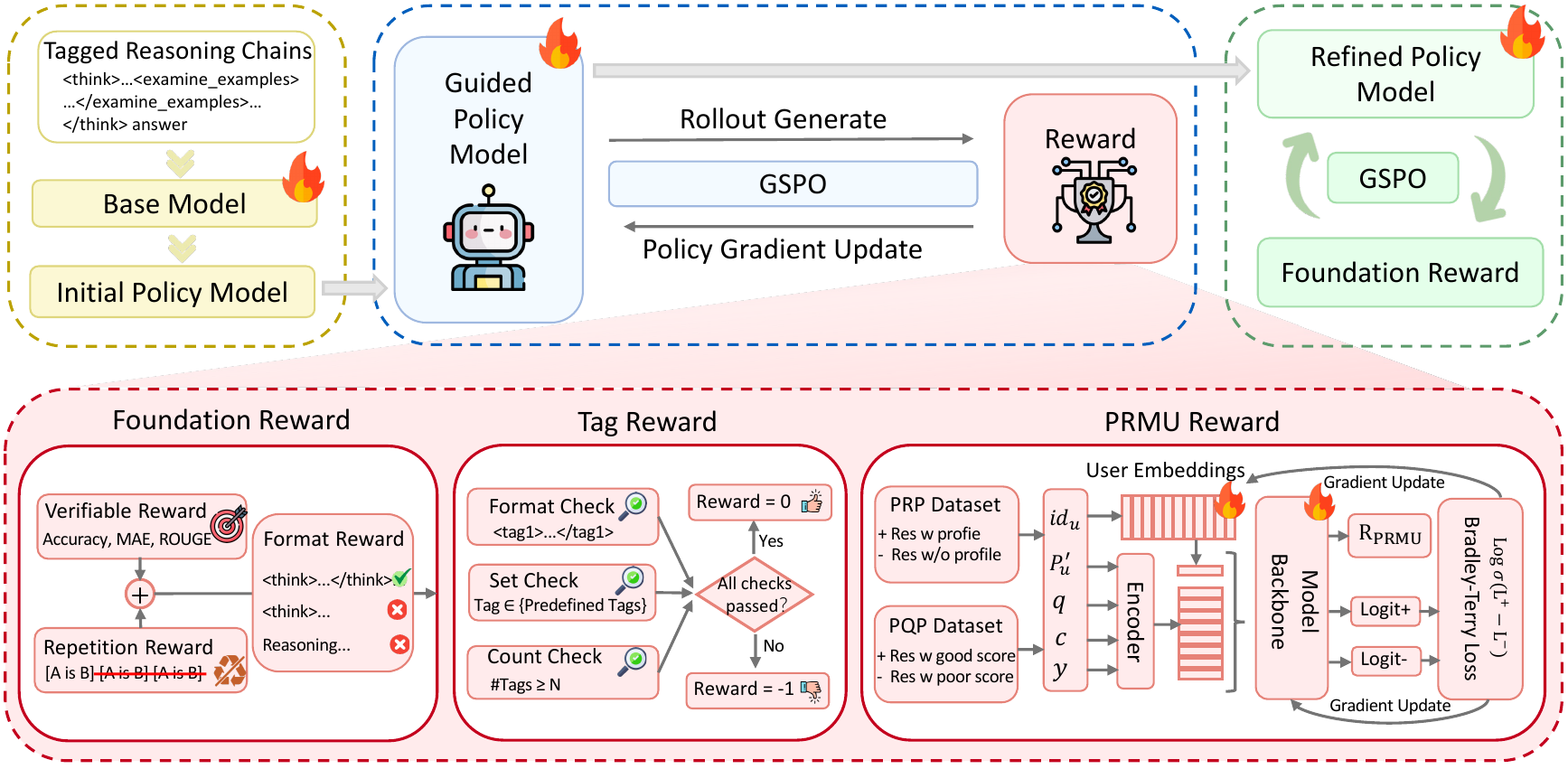}
    \caption{Overview of our proposed multi-stage training framework. An initial policy model is obtained via SFT on tagged reasoning chains. The model is then refined through two sequential RL phases: (1) a \textbf{Guided RL} stage using a complex, multi-component reward (including Tag and PRMU rewards) to learn structured reasoning, and (2) an \textbf{Exploratory RL} stage with a Foundation reward to further boost performance.}
    \label{fig:main}
\end{figure*}

\subsection{Reasoning Structure Initialization}
\label{sec:structure_init}

\noindent \textbf{Tagged Reasoning Chains Construction Pipeline}
Automated construction pipelines are increasingly used to build training resources~\citep{jin2026finrpt}.
To equip the model with a structured reasoning paradigm, we design a multi-stage pipeline to construct a tagged reasoning dataset (Figure~\ref{fig:data}). We randomly sample 1,000 instances from each of the six training tasks in the LaMP dataset~\citep{salemi2024lamp}, employing Qwen3-235B-A22B-Thinking-2507~\citep{yang2025qwen3} to generate 16 candidate reasoning chains per instance via rollout.

To ensure dataset integrity, we applied a rigorous two-stage filter. First, an \textit{accuracy filter} retained only correctly answered samples. Second, an \textit{LLM filter} used GPT-4o~\citep{hurst2024gpt} to score these reasoning chains on logic, factual accuracy, completeness, and conciseness. Only instances scoring >15 were retained for subsequent annotation (Prompt in Appendix~\ref{app:prompt_llm_filter}).

The filtered reasoning chains then underwent a two-phase tagging procedure. In the first phase, \textit{exploratory tagging} used GPT-4o to generate unrestricted tags based on the reasoning content, which were then grouped via K-means clustering~\citep{macqueen1967multivariate} into 9 primary tags: \texttt{<analyze\_input>}, \texttt{<examine\_examples>}, \texttt{<identify\_patterns>}, \texttt{<evaluate\_reference>}, \texttt{<compare\_entities>}, \texttt{<synthesize\_findings>}, \texttt{<make\_decision>}, \texttt{<verify\_conclusion>} and \texttt{<formulate\_conclusion>}. In the second phase, \textit{restricted tagging} constrained GPT-4o to re-annotate all reasoning chains using only these 9 tags for consistency. Detailed tagging prompts are provided in Appendix~\ref{app:prompt_exploratory_tagging} and Appendix~\ref{app:prompt_restricted_tagging}.

\noindent \textbf{Foundational SFT for Syntax Bootstrapping}
Our framework employs a compact but high-quality dataset at the SFT stage, retaining approximately 2,000 instances.
Rather than encoding personalization knowledge, the objective of this stage is to instill the syntactic grammar of structured reasoning, specifically the capacity to produce well-formed tagged thought steps such as \texttt{<analyze\_input>} and \texttt{<identify\_patterns>}. Without this initialization, tag-based process supervision in the subsequent RL stage would be infeasible. We minimize the negative log-likelihood of generating reasoning chain $c$ and answer $y$ given query $q$ and user profile $p_u$:
\begin{equation}
    \mathcal{J}_{\text{SFT}}(\theta) = - \sum_{(q, p_u, c, y) \in \mathcal{D}} \log P_{\theta}(c, y | q, p_u),
\end{equation}
where $\mathcal{D}$ is the labeled dataset and $\theta$ are the model parameters.

\subsection{Tag-Guided Reinforcement Learning}
\label{sec:tag_guided_rl}

\subsubsection{Tag-Based Process Supervision}
\label{sec:tag_reward}

By compelling the model to externalize its logic into discrete, semantically tagged steps, we transform the opaque internal reasoning process into an observable sequence that can be directly supervised. This resolves the degeneracy of outcome-only rewards, which cannot distinguish between a model that carefully analyzes user history and one that ignores it but guesses correctly.

The Tag Reward $R_{\text{tag}}$ enforces both the structural and semantic correctness of the tagged reasoning chain. It operates as a penalty-based signal that applies a set of logical checks to the generated chain $c$:
\begin{equation}
\resizebox{0.89\linewidth}{!}{$
    R_{\text{tag}}(c, y) = \begin{cases}
        0 & \text{if all logical checks pass} \\
        -1 & \text{otherwise}
    \end{cases}.
    $}
\end{equation}
The logical checks include three verifications: (1) \textit{format verification}, ensuring that each tag is syntactically well-formed with correct opening and closing delimiters; (2) \textit{vocabulary verification}, ensuring that every tag used belongs to the predefined set of 9 primary tags established during data construction; and (3) \textit{coverage verification}, ensuring that the reasoning chain meets a minimum tag count, preventing degenerate solutions that satisfy format requirements with a single trivial tag. Together, these checks enforce a cognitive structure on the model's thought process, guiding it along a principled chain of thought.

\subsubsection{Fine-Grained Alignment via PRMU}
\label{sec:prmu}

While tag-based supervision effectively guides the model toward a principled personalization reasoning paradigm, fine-grained alignment with individual user preferences requires an additional user-aware signal. To provide this signal, we introduce the Personalization Reward Model with User Embeddings (PRMU).

Unlike standard reward model architectures, PRMU incorporates learnable user embeddings $E_u$ to capture individual preferences. This architectural modification enables it to provide a granular reward signal that prioritizes reasoning which is not only accurate but also highly tailored to the user's profile.
PRMU is trained on two preference datasets ($\sim$10k samples each). The Profile-Reasoning Preference (PRP) dataset contrasts responses generated with a user profile (preferred) against those generated without (rejected), teaching the model to value profile utilization. The Personalized-Quality Preference (PQP) dataset contains pairs of personalized responses where preference is determined by correctness or ROUGE score, thereby training the model to discern reasoning quality among personalized outputs.

\begin{table*}[h!]
  \centering

  \resizebox{\textwidth}{!}{
    \begin{tabular}{lc|cccccccccccc}
    \toprule
    \multicolumn{2}{l}{\textbf{Dataset $\rightarrow$}} & \multicolumn{2}{c}{\textbf{LaMP-1}} & \multicolumn{2}{c}{\textbf{LaMP-2}} & \multicolumn{2}{c}{\textbf{LaMP-3}} & \multicolumn{2}{c}{\textbf{LaMP-4}} & \multicolumn{2}{c}{\textbf{LaMP-5}} & \multicolumn{2}{c}{\textbf{LaMP-7}} \\
    \cmidrule(lr){1-2}
    \cmidrule(lr){3-4} \cmidrule(lr){5-6} \cmidrule(lr){7-8} \cmidrule(lr){9-10} \cmidrule(lr){11-12} \cmidrule(lr){13-14}
    \textbf{Method} & \multicolumn{1}{c}{\textbf{R}} & \textbf{ACC $\uparrow$} & \textbf{F1 $\uparrow$} & \textbf{ACC $\uparrow$} & \textbf{F1 $\uparrow$} & \textbf{MAE $\downarrow$} & \textbf{RMSE $\downarrow$} & \textbf{R-1 $\uparrow$} & \textbf{R-L $\uparrow$} & \textbf{R-1 $\uparrow$} & \textbf{R-L $\uparrow$} & \textbf{R-1 $\uparrow$} & \textbf{R-L $\uparrow$} \\
    \midrule
    \textit{\textbf{Previous Method}} & \multicolumn{1}{c}{} &       &       &       &       &       &       &       &       &       &       &       &  \\
    \textbf{Zero-shot} &  \ding{55}    & 0.498 & 0.470 & 0.318 & 0.244 & 0.639 & 0.983 & 0.144 & 0.125 & 0.417 & 0.351 & 0.465 & 0.413 \\
    \textbf{Zero-shot-R} &   \ding{51}    & 0.477 & 0.483 & 0.389 & 0.347 & 0.416 & 0.778 & 0.131 & 0.115 & 0.354 & 0.306 & 0.431 & 0.383 \\
    \textbf{RAG} &   \ding{55}     & 0.668 & 0.645 & 0.414 & 0.361 & 0.354 & 0.710 & 0.158 & 0.139 & 0.453 & 0.384 & 0.473 & 0.419 \\
    \textbf{RAG-R (Base)} &  \ding{51}     & 0.717 & 0.722 & 0.453 & 0.413 & 0.291 & 0.645 & 0.152 & 0.137 & 0.434 & 0.365 & 0.439 & 0.391 \\
    \textbf{PAG} &   \ding{55}     & 0.677 & 0.649 & 0.420 & 0.367 & 0.337 & 0.675 & 0.167 & 0.148 & 0.452 & 0.385 & 0.479 & 0.426 \\
    \textbf{PAG-R} &   \ding{51}    & 0.731 & 0.736 & 0.470 & 0.417 & 0.289 & 0.627 & 0.160 & 0.142 & 0.408 & 0.349 & 0.428 & 0.380 \\
    \textbf{SFT} &  \ding{55}      & 0.670 & 0.654 & 0.511 & 0.461 & 0.273 & 0.569 & 0.196 & 0.178 & 0.455 & 0.393 & 0.498 & 0.445 \\
    \textbf{SFT-R} &    \ding{51}   & 0.720 & 0.723 & 0.460 & 0.419 & 0.343 & 0.889 & 0.161 & 0.147 & 0.438 & 0.377 & 0.439 & 0.389 \\
    \textbf{SFT-Ind} &  \ding{55}      & 0.717 & 0.717 & 0.532 & 0.488 & 0.269 & 0.568 & 0.207 & 0.187 & 0.463 & 0.411 & 0.507 & 0.454 \\
    \textbf{SFT-Ind-R} &    \ding{51}   & 0.729 & 0.731 & 0.463 & 0.419 & 0.366 & 1.001 & 0.151 & 0.138 & 0.432 & 0.374 & 0.433 & 0.383 \\
    \textbf{PPlug} &   \ding{55}     & 0.698 & 0.699 & 0.535 & 0.489 & 0.261 & 0.532 & 0.213 & 0.195 & 0.486 & 0.434 & 0.521 & 0.465 \\
    \textbf{HYDRA-Adapter} &  \ding{55}      & 0.692 & 0.692 & 0.482 & 0.455 & 0.320 & 0.663 & 0.159 & 0.138 & 0.457 & 0.395 & 0.483 & 0.423 \\
    \textbf{R2P} &   \ding{51}    & 0.729 & 0.730 & 0.487 & 0.459 & 0.267 & 0.557 & 0.176 & 0.155 & 0.459 & 0.396 & 0.489 & 0.426 \\
    \textbf{PrLM} &   \ding{51}    & 0.731 & 0.731 & 0.534 & 0.504 & 0.288 & 0.635 & 0.183 & 0.169 & 0.499 & 0.438 & 0.513 & 0.459 \\
    \midrule
    \textit{\textbf{State-of-the-Art LLMs}} & \multicolumn{1}{c}{} &       &       &       &       &       &       &       &       &       &       &       &  \\
    \textbf{GPT-4o} &  \ding{55}      & 0.733 & 0.733 & 0.542 & 0.512 & 0.254 & 0.554 & 0.191 & 0.175 & 0.470 & 0.407 & 0.475 & 0.419 \\
    \textbf{Qwen3-235B-A22B} &    \ding{51}   & 0.715 & 0.720 & 0.511 & 0.488 & 0.280 & 0.633 & 0.177 & 0.158 & 0.450 & 0.396 & 0.455 & 0.409 \\
    \textbf{DeepSeek-R1} &   \ding{51}    & 0.740 & 0.744 & 0.522 & 0.488 & 0.280 & 0.644 & 0.181 & 0.166 & 0.451 & 0.399 & 0.447 & 0.397 \\
    \textbf{Gemini-2.5-Pro} &    \ding{51}   & 0.761 & 0.761 & 0.582 & 0.548 & 0.271 & 0.594 & 0.222 & 0.202 & 0.495 & 0.438 & 0.480 & 0.425 \\
     \textbf{Gemini-3-Pro} & \ding{51} & 0.780 & 0.780 & \underline{0.595} & \underline{0.560} & 0.260 & 0.565 & \underline{0.230} & \underline{0.210} & 0.505 & 0.445 & 0.494 & 0.436 \\
    \midrule
    \textit{\textbf{Our Method}} & \multicolumn{1}{c}{} &       &       &       &       &       &       &       &       &       &       &       &  \\
    \textbf{TagPR w/o RL} &   \ding{51}    & 0.722 & 0.724 & 0.456 & 0.416 & 0.339 & 0.878 & 0.159 & 0.145 & 0.440 & 0.378 & 0.437 & 0.386 \\
    \textbf{TagPR w/o SFT} &  \ding{51}     & 0.747 & 0.747 & 0.543 & 0.510 & 0.271 & 0.593 & 0.194 & 0.181 & 0.502 & 0.441 & 0.525 & 0.469 \\
    \textbf{TagPR w/o Tagging} &   \ding{51}    & 0.748 & 0.749 & 0.546 & 0.511 & 0.264 & 0.447 & 0.199 & 0.185 & 0.518 & 0.448 & 0.528 & 0.471 \\
    \textbf{TagPR w/o Tag-R} &   \ding{51}    & 0.749 & 0.749 & 0.545 & 0.511 & 0.272 & 0.595 & 0.197 & 0.183 & 0.506 & 0.441 & 0.524 & 0.469 \\
    \textbf{TagPR w/o PRMU} &   \ding{51}    & 0.768 & 0.769 & 0.557 & 0.514 & 0.246 & 0.393 & 0.205 & 0.190 & 0.522 & 0.453 & 0.545 & 0.490 \\
    \textbf{TagPR w/o Exp-RL} &   \ding{51}    & \underline{0.789} & \underline{0.789} & 0.582 & 0.543 & \underline{0.233} & \underline{0.302} & 0.224 & 0.204 & \underline{0.536} & \underline{0.467} & \underline{0.560} & \underline{0.501} \\
    \rowcolor{aliceblue!60} \textbf{TagPR} &  \ding{51}     & \textbf{0.803} & \textbf{0.803} & \textbf{0.598} & \textbf{0.557} & \textbf{0.218} & \textbf{0.263} & \textbf{0.234} & \textbf{0.213} & \textbf{0.542} & \textbf{0.471} & \textbf{0.565} & \textbf{0.507} \\
    \bottomrule
    \end{tabular}%
    }
    \caption{Main results on the LaMP benchmark, comparing TagPR against a wide range of baselines, including previous methods, state-of-the-art LLMs, and our own ablation studies. 
  \textbf{Bold} indicates the best performance, and \underline{underline} indicates the second-best. The ``R'' column denotes whether a reasoning step is used (\ding{51}).}
    \label{tab:main}%
\end{table*}%

Initialized from Skywork-Reward-V2-Qwen3-0.6B~\citep{liu2026skywork}, our PRMU architecture first maps a user ID $id_u$ to its corresponding embedding $E_u$. This embedding, along with the query, profile, and reasoning chain, is processed to produce a scalar logit. Both the base reward model parameters $\theta_r$ and the user embeddings $E$ are jointly optimized by minimizing the Bradley-Terry~\citep{bradley1952rank} preference loss:
\begin{equation}
\resizebox{0.89\linewidth}{!}{$
\mathcal{J}(\theta_r, E) = -\mathbb{E}_{(x^{+}, x^{-}) \sim \mathcal{D}} \Big[ \log \sigma \Big( f_{\text{PRMU}}(x^{+}) - f_{\text{PRMU}}(x^{-}) \Big) \Big],
$}
\label{eq:loss_prmu}
\end{equation}
where $x^{+}$ and $x^{-}$ represent the preferred and rejected input tuples from our preference dataset $\mathcal{D}$. The final normalized reward score $R_{\text{PRMU}}$ for the RL phase is obtained via a sigmoid function:
\begin{equation}
R_{\text{PRMU}} = \sigma(f_{\text{PRMU}}(id_u, q, p_u, c, y | E_u; \theta_r)).
\label{eq:reward_prmu}
\end{equation}

\subsubsection{Multi-Stage Policy Optimization}
\label{sec:policy_opt}

With the process supervision signal from $R_{\text{tag}}$ and the user-aware signal from $R_{\text{PRMU}}$ established, we now describe how these components are integrated into a complete multi-stage policy optimization strategy, as illustrated in Figure~\ref{fig:main}.

\noindent \textbf{Composite Reward Design}
The full reward function used in the first RL stage combines five distinct signals. Beyond $R_{\text{tag}}$ and $R_{\text{PRMU}}$, we define three additional components. The Verifiable Reward $R_v$ measures the factual correctness of the response $y$ against a ground-truth reference $y^*$:
\begin{equation}
\resizebox{0.89\linewidth}{!}{$
    R_v(y, y^*) = \begin{cases}
        \text{Accuracy}(y, y^*) & \text{classification} \\
        \text{ROUGE}(y, y^*) & \text{generation}
    \end{cases}.
    $}
\end{equation}
The Format Reward $R_f$ provides a binary signal to enforce the overall structural integrity of the output:
\begin{equation}
\resizebox{0.89\linewidth}{!}{$
    R_f(c, y) = \begin{cases}
        1 & \text{if } c, y \text{ match expected format} \\
        0 & \text{otherwise}
    \end{cases}.
    $}
\end{equation}
The Repetition Reward $R_{\text{rep}}$ penalizes textual redundancy to improve fluency:
\begin{equation}
    R_{\text{rep}}(c, y) = - \frac{|T_n(c, y)| - |U_n(c, y)|}{|T_n(c, y)| + \delta},
\end{equation}
where $T_n$ and $U_n$ are the multiset and set of n-grams in the generation respectively, and $\delta$ is a small constant for numerical stability.

These components are integrated into the composite reward $R$ as follows:
\begin{equation}
    R = \alpha \cdot (R_{v} + R_{\text{rep}}) \cdot R_{f} + \beta \cdot R_{\text{tag}} + \gamma \cdot R_{\text{PRMU}},
\end{equation}
where we set the balancing hyperparameters $\alpha = \beta = 0.8$ and $\gamma = 0.2$. These values deliberately prioritize logical correctness over personalization to prevent factuality degradation and reward hacking. We adopt a multiplicative form for $R_f$ to condition other signals on format compliance, ensuring that rewards are only granted for structurally valid outputs. A sensitivity analysis in Appendix~\ref{app:hyperparameter} confirms robustness across a reasonable range of hyperparameter variations.

\noindent \textbf{Guided RL for Personalization Reasoning}
In the first RL stage, the SFT-initialized model is refined using the full composite reward $R$. This stage teaches the model to simultaneously satisfy structural constraints, factual correctness, and user-specific alignment. For policy optimization, we utilize the GSPO algorithm~\citep{zheng2025groupsequencepolicyoptimization}, which offers greater training stability by operating at the sequence level. The GSPO objective is:
\begin{equation}
\resizebox{0.86\linewidth}{!}{$
    \begin{aligned}
        \mathcal{J}_{\text{GSPO}}(\theta) = \mathbb{E}_{\substack{q \sim \mathcal{D} \\ \{c_i, y_i\}_{i=1}^G \sim  \pi_{\theta_{\text{old}}}(\cdot|q)}} \bigg[ \frac{1}{G} \sum_{i=1}^{G} \min \Big(  s_i(\theta) \hat{A}_i, \\
        \text{clip}(s_i(\theta), 1-\epsilon, 1+\epsilon) \hat{A}_i \Big) \bigg],
    \end{aligned}
$}
\end{equation}
where $s_i(\theta)$ is the sequence-level importance sampling ratio and $\hat{A}_{i}$ is the standardized advantage for each response in a generated group of size $G$.

\noindent \textbf{Exploratory RL for Performance Refinement}
In the final stage, we address performance plateaus that arise from the constraints imposed by the composite reward. Since the model has internalized the reasoning structure, we drop the scaffolding of $R_{\text{tag}}$ and $R_{\text{PRMU}}$ and employ a simplified Foundation Reward targeting fundamental quality metrics:
\begin{equation}
    R_{\text{foundation}} = (R_{v} + R_{\text{rep}}) \cdot R_{f}.
\end{equation}
Removing these constraints allows the model to explore the policy space more freely, optimizing core performance without structural penalties. Ablation studies confirm the superiority of this two-stage design over a single continuous RL stage.

\section{Experiments} 

\subsection{Experimental Setup}

\textbf{Implementation Details} We employ Qwen3-8B as our base model. Our training process consists of SFT on the dataset described in Section~\ref{sec:structure_init}, followed by a two-stage RL phase using data sampled from the LaMP training set. We evaluate our model on the LaMP benchmark, a standard for assessing personalization, reporting results on its validation set as the test set is not public.

\noindent \textbf{Baselines} We conduct a comprehensive comparison against a wide spectrum of baselines. These include: (1) standard methodologies such as Zero-shot, RAG, PAG~\citep{richardson2023integrating}, SFT, SFT-Ind, and their reasoning-enhanced variants (-R); (2) advanced personalization (PPlug~\citep{liu2025llms+}, HYDRA-Adapter~\citep{zhuang2024hydra}) and reasoning-focused techniques (R2P~\citep{luo2025reasoning}, PrLM~\citep{zhang2025prlm}); and (3) state-of-the-art large language models like Gemini-2.5-Pro~\citep{comanici2025gemini}. One primary baseline is the RAG-R (Base) method, which shares our configuration with the original Qwen3-8B model.

More detailed descriptions of all baselines, hyperparameters, benchmark, evaluation metrics, and experimental configurations are in the Appendix~\ref{app:experimental_setup}.

\subsection{Main Results}

The results, presented in Table~\ref{tab:main}, demonstrate that \textbf{TagPR} establishes a new state-of-the-art across all six tasks of the LaMP benchmark. It consistently outperforms a 
suite of baselines, including prior personalization methods, reasoning-focused models, and
larger proprietary LLMs used off-the-shelf without task-specific training.

To isolate the efficacy of our framework, we first conduct an ablation study comparing \textbf{TagPR} against a RAG-R (Base) method. This baseline shares an identical configuration but utilizes the original Qwen3-8B model. The performance gains are substantial: \textbf{TagPR} achieves a 55.5\% relative improvement in ROUGE-L on the LaMP-4 generation task, boosts the F1-score by 34.9\%  on the challenging LaMP-2 classification task, and reduces the MAE by 25.1\% on the LaMP-3 task. These results underscore that our synergistic training paradigm significantly enhances the model's personalization reasoning capabilities.

Notably, our fine-tuned 8B parameter model 
outperforms larger proprietary models evaluated without task-specific training. For instance, on the LaMP-1 task, \textbf{TagPR}'s accuracy of 0.803 surpasses both Gemini-3-Pro (0.780) and GPT-4o (0.733). 

\subsection{Ablation Study}

To dissect the contribution of each component within our framework, we conducted a comprehensive ablation study, with results presented in Table~\ref{tab:main}. Our analysis reveals a clear incremental contribution from each module.

The results first highlight the critical roles of the foundational training stages. The initial SFT phase is essential for bootstrapping the model with our tagged reasoning syntax, and its removal (TagPR w/o SFT) leads to a noticeable performance drop. Building upon this, the multi-stage RL is vital for refining this structure into high-quality personalized logic, and the SFT-only model (TagPR w/o RL) leaves a considerable gap to the full model.
Comparable results between TagPR w/o RL and SFT-R indicate that tags yield no measurable gain during SFT alone, but become effective as process supervision during RL.

Furthermore, our novel reward signals prove effective. The PRMU reward provides a crucial user-aware signal, and its removal (TagPR w/o PRMU) leads to a consistent decline across all tasks. Regarding the tag-based reward, note that TagPR w/o Tagging replaces the 9 personalization-specific tags with generic unstructured reasoning, while TagPR w/o Tag-R retains the tags but removes their associated reward. Both variants underperform the full model, confirming that the semantic content of the personalization-specific tags and their process supervision together drive consistent gains.
Finally, our two-stage training design is validated as superior to a single, continuous RL stage (TagPR w/o Exp-RL). Collectively, these findings affirm that the synergistic integration of each carefully designed component is the key to TagPR's success.

\subsection{Generalization Assessment}

To evaluate whether \textbf{TagPR} learns generalizable personalization capabilities,
we assess its zero-shot generalization performance on two other public benchmarks and  a new self-constructed benchmark. 

We constructed a benchmark from Dianping\footnote{\url{https://www.dianping.com/}.}, a prominent Chinese user-generated content platform. This setup poses a stringent test involving unseen domains, task formats, and a different language. A more detailed benchmark introduction is provided in the Appendix~\ref{app:dianping_concise}. As shown in Table~\ref{tab:tab_generalization}, \textbf{TagPR} demonstrates exceptional generalization capabilities. It achieves state-of-the-art results across the benchmark, securing the top score on the majority of metrics 
and leading proprietary models like GPT-4o evaluated without task-specific training. 
\textbf{TagPR} creates a highly generalizable personalization reasoning model effective across diverse domains, tasks, and languages.

\begin{table}[htbp]
  \centering
  \resizebox{\linewidth}{!}{%
    \begin{tabular}{lcccccc}
    \toprule
    \textbf{Dataset $\rightarrow$} & \multicolumn{2}{c}{\textbf{Dianping-Content}} & \multicolumn{2}{c}{\textbf{Dianping-Title}} & \multicolumn{2}{c}{\textbf{Dianping-Paraph}} \\
    \cmidrule(lr){1-1}
    \cmidrule(lr){2-3} \cmidrule(lr){4-5} \cmidrule(lr){6-7}
    \textbf{Method} & \textbf{R-1 $\uparrow$} & \textbf{R-L $\uparrow$} & \textbf{R-1 $\uparrow$} & \textbf{R-L $\uparrow$} & \textbf{R-1 $\uparrow$} & \textbf{R-L $\uparrow$} \\
    \midrule
    RAG & 0.200  & 0.151  & 0.209  & 0.184  & 0.598  & 0.568  \\
    RAG-R (Base) & 0.183  & 0.144  & 0.197  & 0.173  & 0.517  & 0.461  \\
    SFT & 0.189  & 0.123  & 0.228  & 0.210  & 0.603  & 0.571  \\
    SFT-R & 0.187  & 0.145  & 0.198  & 0.177  & 0.498  & 0.423  \\
    GPT-4o & 0.207  & 0.168  & \underline{0.236}  & \underline{0.211}  & \underline{0.606}  & \underline{0.573}  \\
    Gemini-2.5-Pro & \textbf{0.217}  & \underline{0.170}  & 0.215  & 0.195  & 0.564  & 0.475  \\
    \rowcolor{aliceblue!60} \textbf{TagPR} & \underline{0.216}  & \textbf{0.171}  & \textbf{0.240}  & \textbf{0.218}  & \textbf{0.617}  & \textbf{0.583}  \\
    \bottomrule
    \end{tabular}%
    }
      \caption{Zero-shot cross-lingual generalization performance on the three Dianping datasets. 
  The best results are in \textbf{bold}, and the second-best are \underline{underlined}. Our TagPR demonstrates superior performance.}
  \label{tab:tab_generalization}
\end{table}

To further assess the generalization capabilities of TagPR, we evaluated the model on partial test sets of LongLaMP~\citep{kumar2024longlamp} and PGraphRAG~\citep{au2025personalized} datasets in the Appendix~\ref{app:generalization_assessment}. 
These benchmarks feature ultra-long user histories and sparse user profiles, respectively, providing complementary tests of personalization reasoning under challenging conditions.
As shown in Table~\ref{tab:tab_generalization_long_pgarph} in Appendix, our model demonstrated superior performance, verifying the robust generalization of its learned personalized reasoning capability. 

\subsection{Robustness Assessment}

\begin{figure*}
 \centering
\includegraphics[width=1.0\linewidth]{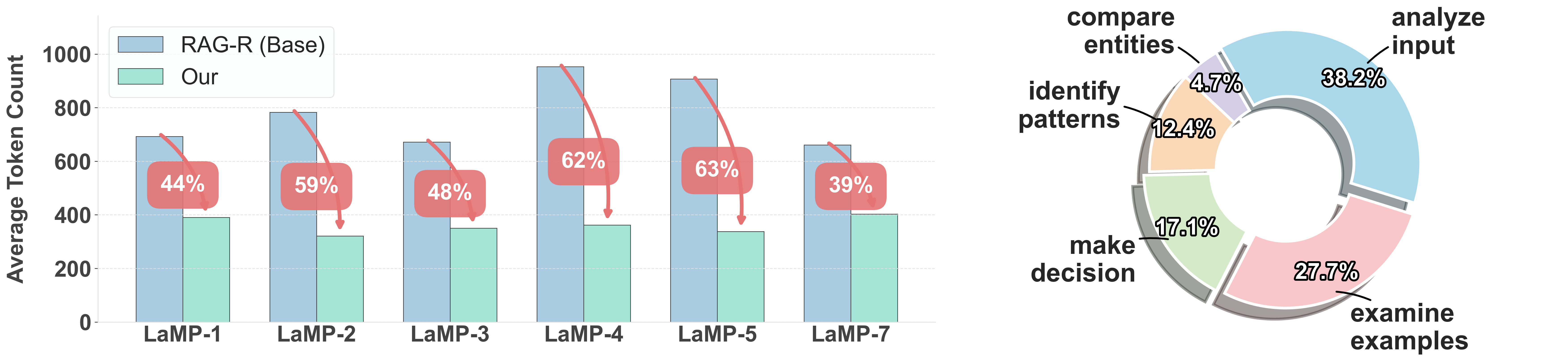}
\caption{
\textbf{Left}: Comparison of reasoning chain length
between TagPR and Base on the LaMP validation set.
Our model achieves substantial length reductions across all tasks, indicating greater efficiency. 
\textbf{Right}: Frequency distribution of the five core reasoning tags generated by our model. 
}
\label{fig:reasoning_analysis}
\end{figure*}

We evaluate the robustness of \textbf{TagPR} against baselines SFT and Base by varying two key factors: user profile length and profile retrieval method. Figure~\ref{fig:roubustness_assessment} in Appendix~\ref{app:robustness} presents the results on the representative LaMP-2 and LaMP-4 tasks.

First, we analyze the effect of profile length by varying the number of historical interactions from 2 to 32. The top row of Figure~\ref{fig:roubustness_assessment} shows that \textbf{TagPR} consistently outperforms the baselines across all lengths. Notably, TagPR's performance improves rapidly and starts to plateau with just 8 interactions, indicating its high data efficiency in distilling user preferences. In contrast, the baselines show more gradual improvements and maintain a significant performance gap.

Second, we assess the model's sensitivity to the profile retrieval method. We compare our default Recency-based retriever with three alternatives: a sparse retriever (BM25~\cite{robertson2009probabilistic}), a dense retriever (Contriever~\cite{izacard2021unsupervised}), and Random selection. As shown in the bottom row, \textbf{TagPR} demonstrates remarkable stability and maintains its superior performance across all retrieval strategies. Even with randomly selected profiles, TagPR's performance degradation is minimal, suggesting its reasoning process can effectively identify and utilize relevant information regardless of the profile quality. 

\subsection{Further Analysis}
This section validates the PRMU design and analyzes tagged reasoning chains, with further case studies and semantic consistency verification in Appendix~\ref{app:add_further_analysis}.

\subsubsection{Personalization Reward Model Design}

To validate our proposed PRMU, we conducted a comprehensive ablation study to assess the contribution of its core components. Table~\ref{tab:ablation_prmu} compares four configurations: our full PRMU, PRMU without user embeddings (w/o UE), a baseline using an untrained reward model (Untrained RM), and a baseline with no reward model (w/o RM). Our findings first reveal that employing an off-the-shelf
reward model offers no consistent advantage over having no reward model at all. In fact, it proved detrimental in certain cases, yielding a noisy and misaligned signal. 
This underscores the necessity of fine-tuning the reward model on domain-specific data.
Next, training the reward model on our personalization dataset, even without user-specific information (PRMU w/o UE), yields 
improvements across all metrics. 
The most significant performance gains are realized with the full PRMU model. By integrating user embeddings to provide a user-aware reward, PRMU consistently outperforms all other variants.

\begin{table*}[htbp]
  \centering
  \resizebox{0.95\textwidth}{!}{
    \begin{tabular}{lcccccccccccc}
    \toprule
    \textbf{Dataset $\rightarrow$ } & \multicolumn{2}{c}{\textbf{LaMP-1}} & \multicolumn{2}{c}{\textbf{LaMP-2}} & \multicolumn{2}{c}{\textbf{LaMP-3}} & \multicolumn{2}{c}{\textbf{LaMP-4}} & \multicolumn{2}{c}{\textbf{LaMP-5}} & \multicolumn{2}{c}{\textbf{LaMP-7}} \\
    \cmidrule(lr){1-1}
    \cmidrule(lr){2-3} \cmidrule(lr){4-5} \cmidrule(lr){6-7} \cmidrule(lr){8-9}
    \cmidrule(lr){10-11} \cmidrule(lr){12-13}
    \textbf{Method} & \textbf{ACC $\uparrow$} & \textbf{F1 $\uparrow$} & \textbf{ACC $\uparrow$} & \textbf{F1 $\uparrow$} & \textbf{MAE $\downarrow$} & \textbf{RMSE $\downarrow$} & \textbf{R-1 $\uparrow$} & \textbf{R-L $\uparrow$} & \textbf{R-1 $\uparrow$} & \textbf{R-L $\uparrow$} & \textbf{R-1 $\uparrow$} & \textbf{R-L $\uparrow$} \\
    \midrule
    \textbf{w/o RM} & 0.768  & 0.769  & 0.557  & 0.514  & 0.246  & 0.393  & 0.205  & 0.190  & 0.522  & 0.453  & 0.545  & 0.490  \\
    \textbf{Untrained RM} & 0.771  & 0.772  & 0.533  & 0.495  & 0.246  & 0.361  & 0.207  & 0.195  & \underline{0.536}  & 0.459  & 0.545  & 0.487  \\
    \textbf{PRMU w/o UE} & \underline{0.784}  & \underline{0.784}  & \underline{0.581}  & \underline{0.541}  & \underline{0.231}  & \underline{0.299}  & \underline{0.215}  & \underline{0.197}  & \underline{0.536}  & \underline{0.467}  & \underline{0.558}  & \underline{0.501}  \\
    \rowcolor{aliceblue!60} \textbf{PRMU} & \textbf{0.803}  & \textbf{0.803}  & \textbf{0.598}  & \textbf{0.557}  & \textbf{0.218}  & \textbf{0.263}  & \textbf{0.234}  & \textbf{0.213}  & \textbf{0.542}  & \textbf{0.471}  & \textbf{0.565}  & \textbf{0.507}  \\
    \bottomrule
    \end{tabular}%
    }
  \caption{Ablation study of PRMU components across LaMP benchmarks. Best and second-best results are in \textbf{bold} and \underline{underlined}.}
  \label{tab:ablation_prmu}%
\end{table*}%

\subsubsection{Tagged Reasoning Chains Analysis}

\noindent \textbf{Reasoning Length}
To assess reasoning efficiency, we compare the average token count of reasoning chains generated by our trained model against the  original Qwen3-8B on the LaMP validation set. As illustrated in Figure~\ref{fig:reasoning_analysis} (Left), \textbf{TagPR} consistently produces more concise reasoning chains, achieving an average token reduction of over 50\%. 
For instance, significant reductions of 59\% on LaMP-2 and 62\% on LaMP-4 are observed. 
Unlike the Base model's tendency toward verbose reasoning, our framework guides the model along a direct logical path, effectively pruning irrelevant steps. 
This demonstrates that effective personalization relies on the precision of reasoning, not its length, leading to enhanced computational efficiency.

\noindent \textbf{Reasoning Tags} As shown in Figure~\ref{fig:reasoning_analysis} (Right), the model relies heavily on contextual grounding (\texttt{<analyze\_input>} 38.2\%, \texttt{<examine\_examples>} 27.7\%). Synthesis and decision tags (\texttt{<identify\_patterns>} 12.4\%, \texttt{<compare\_entities>} 4.7\%, \texttt{<make\_decision>} 17.1\%) occupy a smaller proportion. This indicates a highly evidence-driven approach, prioritizing input understanding over immediate conclusions.

\subsubsection{Reasoning Content Analysis}

To further investigate the reasoning processes qualitatively, we generated word clouds from the reasoning chains produced by the baseline Qwen3-8B and our \textbf{TagPR} model on the LaMP validation set, as shown in Figure~\ref{fig:wordcloud} in Appendix~\ref{app:add_further_analysis}. The visualization reveals a stark contrast in their reasoning styles.

The word cloud for the baseline model is populated by general, conversational terms such as ``maybe'', ``think'', ``example'', and ``review''. This indicates a descriptive, narrative-style reasoning process, where the model verbalizes a general thought process rather than executing a structured plan. In sharp contrast, the \textbf{TagPR} word cloud prominently features action-oriented keywords like ``examine\_examples'', ``analyze\_input'', ``identify\_patterns'', and ``make\_decision'', which are the core components of the functional tags introduced in our framework. This shift demonstrates that \textbf{TagPR} successfully learns to adopt an explicit, structured, and interpretable reasoning schema. Instead of merely describing its thought process, the model actively executes a sequence of defined logical steps, confirming a more efficient and targeted approach to personalization reasoning.

\section{Conclusion}

In this work, we presented TagPR, a novel framework designed to bridge the fundamental gap between general reasoning and personalization reasoning in LLMs. To achieve this, we developed a synergistic training paradigm starting with a data-driven pipeline for SFT to instill tagged reasoning syntax. This is followed by a multi-stage RL process. The initial guided phase utilizes a composite reward that features tag-based process supervision and our novel Personalization Reward Model with User Embeddings. A subsequent exploratory phase is then applied to refine core performance. Extensive evaluations across LaMP, LongLaMP, PGraphRAG, and a self-constructed benchmark demonstrate that TagPR achieves state-of-the-art performance. Ultimately, our work proves that explicit tag-guided process supervision is an effective strategy for eliciting genuine personalization reasoning in large language models.

\section{Limitations \& Potential Risks}

Despite TagPR's promising results, several limitations and risks remain. Our nine-tag vocabulary is derived from reasoning chains in LaMP and may therefore reflect benchmark-specific reasoning patterns. Although it generalizes in our experiments, its fixed scope may limit the model's flexibility in handling highly complex or open-ended reasoning scenarios that require dynamic structure adaptation. The use of learnable user embeddings and deep personalization introduces potential privacy concerns regarding user data retention and the risk of reinforcing existing user biases or creating echo chambers, necessitating further research into privacy-preserving and alignment-focused personalization.

\section*{Acknowledgments}
This work is supported by Meituan. This work is also supported by the Public Computing Cloud, Renmin University of China and by fund for building worldclass universities (disciplines) of Renmin University of China.

\bibliography{custom}

@inproceedings{salemi2024lamp,
  title={LaMP: When Large Language Models Meet Personalization},
  author={Salemi, Alireza and Mysore, Sheshera and Bendersky, Michael and Zamani, Hamed},
  booktitle={Proceedings of the 62nd Annual Meeting of the Association for Computational Linguistics (Volume 1: Long Papers)},
  pages={7370--7392},
  year={2024}
}

@inproceedings{jin2026finrpt,
  title={Finrpt: Dataset, evaluation system and llm-based multi-agent framework for equity research report generation},
  author={Jin, Song and Li, Shuqi and Zhang, Shukun and Yan, Rui},
  booktitle={Proceedings of the AAAI Conference on Artificial Intelligence},
  volume={40},
  number={1},
  pages={507--515},
  year={2026}
}

@inproceedings{macqueen1967multivariate,
  title={Multivariate observations},
  author={MacQueen, J},
  booktitle={Proceedings ofthe 5th Berkeley symposium on mathematical statisticsand probability},
  volume={1},
  number={1967},
  pages={281--297},
  year={1967},
  organization={University of California press Oakland, CA, USA}
}

@article{yang2025qwen3,
  title={Qwen3 technical report},
  author={Yang, An and Li, Anfeng and Yang, Baosong and Zhang, Beichen and Hui, Binyuan and Zheng, Bo and Yu, Bowen and Gao, Chang and Huang, Chengen and Lv, Chenxu and others},
  journal={arXiv preprint arXiv:2505.09388},
  year={2025}
}

@article{guo2025deepseek,
  title={Deepseek-r1: Incentivizing reasoning capability in llms via reinforcement learning},
  author={Guo, Daya and Yang, Dejian and Zhang, Haowei and Song, Junxiao and Wang, Peiyi and Zhu, Qihao and Xu, Runxin and Zhang, Ruoyu and Ma, Shirong and Bi, Xiao and others},
  journal={arXiv preprint arXiv:2501.12948},
  year={2025}
}

@article{yu2026dapo,
  title={Dapo: An open-source llm reinforcement learning system at scale},
  author={Yu, Qiying and Zhang, Zheng and Zhu, Ruofei and Yuan, Yufeng and Zuo, Xiaochen and Yue, Yu and Dai, Weinan and Fan, Tiantian and Liu, Gaohong and Liu, Lingjun and others},
  journal={Advances in Neural Information Processing Systems},
  volume={38},
  pages={113222--113244},
  year={2026}
}

@misc{zheng2025groupsequencepolicyoptimization,
      title={Group Sequence Policy Optimization},
      author={Chujie Zheng and Shixuan Liu and Mingze Li and Xiong-Hui Chen and Bowen Yu and Chang Gao and Kai Dang and Yuqiong Liu and Rui Men and An Yang and Jingren Zhou and Junyang Lin},
      year={2025},
      eprint={2507.18071},
      archivePrefix={arXiv},
      primaryClass={cs.LG},
      url={https://arxiv.org/abs/2507.18071},
}

@article{kumar2024longlamp,
  title={Longlamp: A benchmark for personalized long-form text generation},
  author={Kumar, Ishita and Viswanathan, Snigdha and Yerra, Sushrita and Salemi, Alireza and Rossi, Ryan A and Dernoncourt, Franck and Deilamsalehy, Hanieh and Chen, Xiang and Zhang, Ruiyi and Agarwal, Shubham and others},
  journal={arXiv preprint arXiv:2407.11016},
  year={2024}
}

@inproceedings{salemi2024optimization,
  title={Optimization methods for personalizing large language models through retrieval augmentation},
  author={Salemi, Alireza and Kallumadi, Surya and Zamani, Hamed},
  booktitle={Proceedings of the 47th International ACM SIGIR Conference on Research and Development in Information Retrieval},
  pages={752--762},
  year={2024}
}

@inproceedings{mysore2024pearl,
  title={Pearl: Personalizing large language model writing assistants with generation-calibrated retrievers},
  author={Mysore, Sheshera and Lu, Zhuoran and Wan, Mengting and Yang, Longqi and Sarrafzadeh, Bahareh and Menezes, Steve and Baghaee, Tina and Gonzalez, Emmanuel Barajas and Neville, Jennifer and Safavi, Tara},
  booktitle={Proceedings of the 1st Workshop on Customizable NLP: Progress and Challenges in Customizing NLP for a Domain, Application, Group, or Individual (CustomNLP4U)},
  pages={198--219},
  year={2024}
}

@article{richardson2023integrating,
  title={Integrating summarization and retrieval for enhanced personalization via large language models},
  author={Richardson, Chris and Zhang, Yao and Gillespie, Kellen and Kar, Sudipta and Singh, Arshdeep and Raeesy, Zeynab and Khan, Omar Zia and Sethy, Abhinav},
  journal={arXiv preprint arXiv:2310.20081},
  year={2023}
}

@article{wu2024understanding,
  title={Understanding the role of user profile in the personalization of large language models},
  author={Wu, Bin and Shi, Zhengyan and Rahmani, Hossein A and Ramineni, Varsha and Yilmaz, Emine},
  journal={arXiv preprint arXiv:2406.17803},
  year={2024}
}

@inproceedings{jin2026diningbench,
  title={Diningbench: A hierarchical multi-view benchmark for perception and reasoning in the dietary domain},
  author={Jin, Song and Zhang, Juntian and Zhang, Xun and Tian, Zeying and Jiang, Fei and Yin, Guojun and Lin, Wei and Liu, Yong and Yan, Rui},
  booktitle={Proceedings of the 64th Annual Meeting of the Association for Computational Linguistics (Volume 1: Long Papers)},
  pages={35289--35310},
  year={2026}
}

@inproceedings{qiu2025measuring,
  title={Measuring what makes you unique: Difference-aware user modeling for enhancing llm personalization},
  author={Qiu, Yilun and Zhao, Xiaoyan and Zhang, Yang and Bai, Yimeng and Wang, Wenjie and Cheng, Hong and Feng, Fuli and Chua, Tat-Seng},
  booktitle={Findings of the Association for Computational Linguistics: ACL 2025},
  pages={21258--21277},
  year={2025}
}

@inproceedings{tan2024democratizing,
  title={Democratizing large language models via personalized parameter-efficient fine-tuning},
  author={Tan, Zhaoxuan and Zeng, Qingkai and Tian, Yijun and Liu, Zheyuan and Yin, Bing and Jiang, Meng},
  booktitle={Proceedings of the 2024 Conference on Empirical Methods in Natural Language Processing},
  pages={6476--6491},
  year={2024}
}

@inproceedings{tan2024personalized,
  title={Personalized Pieces: Efficient Personalized Large Language Models through Collaborative Efforts},
  author={Tan, Zhaoxuan and Liu, Zheyuan and Jiang, Meng},
  booktitle={Proceedings of the 2024 Conference on Empirical Methods in Natural Language Processing},
  pages={6459--6475},
  year={2024}
}

@inproceedings{liu2025llms+,
  title={Llms+ persona-plug= personalized llms},
  author={Liu, Jiongnan and Zhu, Yutao and Wang, Shuting and Wei, Xiaochi and Min, Erxue and Lu, Yu and Wang, Shuaiqiang and Yin, Dawei and Dou, Zhicheng},
  booktitle={Proceedings of the 63rd Annual Meeting of the Association for Computational Linguistics (Volume 1: Long Papers)},
  pages={9373--9385},
  year={2025}
}

@article{zhuang2024hydra,
  title={Hydra: Model factorization framework for black-box llm personalization},
  author={Zhuang, Yuchen and Sun, Haotian and Yu, Yue and Qiang, Rushi and Wang, Qifan and Zhang, Chao and Dai, Bo},
  journal={Advances in Neural Information Processing Systems},
  volume={37},
  pages={100783--100815},
  year={2024}
}

@article{li2023teach,
  title={Teach LLMs to Personalize--An Approach inspired by Writing Education},
  author={Li, Cheng and Zhang, Mingyang and Mei, Qiaozhu and Wang, Yaqing and Hombaiah, Spurthi Amba and Liang, Yi and Bendersky, Michael},
  journal={arXiv preprint arXiv:2308.07968},
  year={2023}
}

@article{jang2023personalized,
  title={Personalized soups: Personalized large language model alignment via post-hoc parameter merging},
  author={Jang, Joel and Kim, Seungone and Lin, Bill Yuchen and Wang, Yizhong and Hessel, Jack and Zettlemoyer, Luke and Hajishirzi, Hannaneh and Choi, Yejin and Ammanabrolu, Prithviraj},
  journal={arXiv preprint arXiv:2310.11564},
  year={2023}
}

@inproceedings{kim2026rpm,
  title={RPM: Reasoning-level personalization for black-box large language models},
  author={Kim, Jieyong and Kim, Tongyoung and Yoon, Soojin and Kim, Jaehyung and Lee, Dongha},
  booktitle={International Conference on Learning Representations},
  volume={2026},
  pages={84214--84251},
  year={2026}
}

@article{luo2025reasoning,
  title={Reasoning Meets Personalization: Unleashing the Potential of Large Reasoning Model for Personalized Generation},
  author={Luo, Sichun and Deng, Guanzhi and Xu, Jian and Zhang, Xiaojie and Hou, Hanxu and Song, Linqi},
  journal={arXiv preprint arXiv:2505.17571},
  year={2025}
}

@inproceedings{zhang2026viper,
  title={ViPER: Empowering the Self-Evolution of Visual Perception Abilities in Vision-Language Models},
  author={Zhang, Juntian and Jin, Song and Cheng, Chuanqi and Liu, Yuhan and Lin, Yankai and Zhang, Xun and Jiang, Fei and Yin, Guojun and Lin, Wei and Yan, Rui and others},
  booktitle={International Conference on Learning Representations},
  volume={2026},
  pages={98012--98039},
  year={2026}
}

@article{salemi2025reasoning,
  title={Reasoning-Enhanced Self-Training for Long-Form Personalized Text Generation},
  author={Salemi, Alireza and Li, Cheng and Zhang, Mingyang and Mei, Qiaozhu and Kong, Weize and Chen, Tao and Li, Zhuowan and Bendersky, Michael and Zamani, Hamed},
  journal={arXiv preprint arXiv:2501.04167},
  year={2025}
}

@article{li2025extended,
  title={Extended Inductive Reasoning for Personalized Preference Inference from Behavioral Signals},
  author={Li, Jia-Nan and Guan, Jian and Wu, Wei and Yan, Rui},
  journal={arXiv preprint arXiv:2505.18071},
  year={2025}
}

@inproceedings{zhang2025prlm,
  title={Prlm: Learning explicit reasoning for personalized rag via contrastive reward optimization},
  author={Zhang, Kepu and Shi, Teng and Yu, Weijie and Xu, Jun},
  booktitle={Proceedings of the 34th ACM International Conference on Information and Knowledge Management},
  pages={5484--5488},
  year={2025}
}

@inproceedings{liu2026skywork,
  title={Skywork-reward-v2: Scaling preference data curation via human-ai synergy},
  author={Liu, Yuhao and Zeng, Liang and Xiao, Yuzhen and He, Jujie and Liu, Jiacai and Wang, Chaojie and Yan, Rui and Shen, Wei and Zhang, Fuxiang and Xu, Jiacheng and others},
  booktitle={International Conference on Learning Representations},
  volume={2026},
  pages={133805--133838},
  year={2026}
}

@article{bradley1952rank,
  title={Rank analysis of incomplete block designs: I. the method of paired comparisons},
  author={Bradley, Ralph Allan and Terry, Milton E},
  journal={Biometrika},
  volume={39},
  number={3/4},
  pages={324--345},
  year={1952},
  publisher={JSTOR}
}

@article{hurst2024gpt,
  title={Gpt-4o system card},
  author={Hurst, Aaron and Lerer, Adam and Goucher, Adam P and Perelman, Adam and Ramesh, Aditya and Clark, Aidan and Ostrow, AJ and Welihinda, Akila and Hayes, Alan and Radford, Alec and others},
  journal={arXiv preprint arXiv:2410.21276},
  year={2024}
}

@article{comanici2025gemini,
  title={Gemini 2.5: Pushing the frontier with advanced reasoning, multimodality, long context, and next generation agentic capabilities},
  author={Comanici, Gheorghe and Bieber, Eric and Schaekermann, Mike and Pasupat, Ice and Sachdeva, Noveen and Dhillon, Inderjit and Blistein, Marcel and Ram, Ori and Zhang, Dan and Rosen, Evan and others},
  journal={arXiv preprint arXiv:2507.06261},
  year={2025}
}

@inproceedings{au2025personalized,
  title={Personalized graph-based retrieval for large language models},
  author={Au, Steven and Dimacali, Cameron and Pedirappagari, Ojasmitha and Park, Namyong and Dernoncourt, Franck and Wang, Yu and Kanakaris, Nikos and Deilamsalehy, Hanieh and Rossi, Ryan A and Ahmed, Nesreen K},
  booktitle={Proceedings of the 39th Pacific Asia Conference on Language, Information and Computation},
  pages={678--698},
  year={2025}
}

@article{robertson2009probabilistic,
  title={The probabilistic relevance framework: BM25 and beyond},
  author={Robertson, Stephen and Zaragoza, Hugo},
  journal={Foundations and trends{\textregistered} in information retrieval},
  volume={4},
  number={1-2},
  pages={1--174},
  year={2009},
  publisher={Emerald Publishing Limited}
}

@article{izacard2021unsupervised,
  title={Unsupervised dense information retrieval with contrastive learning},
  author={Izacard, Gautier and Caron, Mathilde and Hosseini, Lucas and Riedel, Sebastian and Bojanowski, Piotr and Joulin, Armand and Grave, Edouard},
  journal={arXiv preprint arXiv:2112.09118},
  year={2021}
}

@inproceedings{jin2025beyond,
  title={Beyond static testbeds: An interaction-centric agent simulation platform for dynamic recommender systems},
  author={Jin, Song and Zhang, Juntian and Liu, Yuhan and Zhang, Xun and Zhang, Yufei and Yin, Guojun and Jiang, Fei and Lin, Wei and Yan, Rui},
  booktitle={Proceedings of the 2025 Conference on Empirical Methods in Natural Language Processing},
  pages={18914--18931},
  year={2025}
}

@article{qiu2026preference,
  title={Preference-Aware Rubric Learning for Personalized Evaluation},
  author={Qiu, Yilun and Zhao, Xiaoyan and Zhang, Yang and Chen, Yuxin and Yan, Cilin and Cai, Jiayin and Jiang, Xiaolong and Hu, Yao and Yamakata, Yoko and Chua, Tat-Seng},
  journal={arXiv preprint arXiv:2605.31545},
  year={2026}
}

\clearpage

\appendix

\section{Detailed Experimental Setup}
\label{app:experimental_setup}

This section provides a detailed description of our experimental setup, including implementation details, benchmark information, and baseline configurations.

\subsection{Implementation Details}
\textbf{Backbone Model} We use Qwen3-8B~\citep{yang2025qwen3} as our base model for all experiments unless otherwise specified.

\noindent \textbf{Supervised Fine-Tuning (SFT)} The SFT stage was conducted on 8 A100 GPUs. We used a learning rate of 1e-5 and a global batch size of 64. The model was trained for 2 epochs on the dataset described in Section~\ref{sec:structure_init}.

\noindent \textbf{Reinforcement Learning (RL)}
Data Sampling: We sampled data from the LaMP training set for RL. Specifically, we randomly sampled 1,024 examples for each of the LaMP-1, LaMP-3, LaMP-4, LaMP-5, and LaMP-7 tasks. For the more challenging LaMP-2 task, we sampled 3,200 examples. Training Parameters: The first RL stage was trained for 13 epochs, and the second stage was trained for 2 epochs. Both stages were conducted on 8 A100 GPUs with a global batch size of 128 and a learning rate of 1e-6. Policy Rollout: During the policy rollout stage, we set the temperature to 1.0 and top-p to 1.0, generating 5 responses for each prompt. Other Hyperparameters: The low and high clip ratios for the GSPO algorithm were set to 0.0003 and 0.0004, respectively. For the repetition penalty reward, we used n-grams of size 4. For the tag reward, the minimum required number of tags was set to 3.

\subsection{Benchmark Details}
\textbf{Dataset} We use the LaMP benchmark, a widely-adopted benchmark for evaluating the personalization capabilities of LLMs. It requires models to analyze user historical profiles to answer current queries. Since the official test set is not publicly available, all our evaluations are conducted on the official validation set. LaMP-6 was excluded from our evaluation due to its unavailability. We evaluated on the complete validation dataset for all other tasks. The detailed data statistics of LaMP are shown in Table~\ref{tab:lamp_stat}.

\begin{table}[!ht]
 \center
 
 \small
  \resizebox{\linewidth}{!}{
  \begin{tabular}{llrrr}
  	\toprule
        {\textbf{Task}} & {\textbf{Task Type}} & \textbf{\#Train} & \textbf{\#Val} & \textbf{\#Classes} \\
        \midrule
        LaMP-1 & Binary classification & 6,542 & 1,500 & 2 \\
        LaMP-2 & Categorical classification & 5,073 & 1,410 & 15 \\
        LaMP-3 & Ordinal classification & 20,000 & 2,500 & 5 \\
        LaMP-4 & Text generation & 12,500 & 1,500 & - \\
        LaMP-5 & Text generation & 14,682 & 1,500 & - \\    
        LaMP-7 & Text generation & 13,437 & 1,498 & - \\
 \bottomrule
  \end{tabular}
  }
   \caption{Data statistics of the LaMP benchmark.} 
  \label{tab:lamp_stat}
\end{table}

\noindent \textbf{Evaluation Metrics} Following the original LaMP benchmark, we employ the following metrics:

Classification Tasks (LaMP-1 \& LaMP-2). Task Descriptions: LaMP-1 (Personalized Citation): A binary classification task where the model determines which of two candidate references an author is likely to cite based on their profile. LaMP-2 (Personalized Movie Tagging): A multi-class classification task requiring the model to select the most appropriate tag from 15 candidates for a movie, conditioned on the user’s tagging history. Metrics \& Calculation: We report Accuracy and F1-Score. For Accuracy, we calculate the ratio of correctly predicted samples to the total samples. For F1-Score: In LaMP-1 (binary), we report the standard F1 for the positive class. In LaMP-2 (multi-class), we follow the benchmark standard and report Macro-F1.

Rating Task (LaMP-3). Task Description: LaMP-3 (Personalized Product Rating) is an ordinal classification task. The model predicts a specific integer rating (1–5 stars) for a product based on the user's historical reviews. Metrics: We employ MAE (Mean Absolute Error) and RMSE (Root Mean Square Error). These metrics are standard for quantifying the deviation between predicted scores and ground-truth user ratings.

Text Generation Tasks (LaMP-4, 5, 7). Task Descriptions: These tasks involve generating text tailored to a user's specific style: LaMP-4: Personalized news headline generation. LaMP-5: Personalized scholarly title generation. LaMP-7: Personalized tweet paraphrasing. Metrics: We utilize ROUGE-1 and ROUGE-L. ROUGE-1: measures the overlap of unigrams to assess content adequacy. ROUGE-L: is based on the Longest Common Subsequence, evaluating fluency and structural similarity between the generated text and the reference.

\subsection{Baselines and Comparison Setup}
To rigorously evaluate our proposed method, we benchmark it against a wide spectrum of baselines. For a fair comparison, all methods are built upon the Qwen3-8B base model and utilize the user's 8 most recent profiles as input, unless specified otherwise (e.g., proprietary models like GPT-4o). 

The baselines are categorized as follows. Standard methodologies include Zero-shot, which generates responses without user profiles as a non-personalized lower bound; standard Retrieval-Augmented Generation (RAG); Personalization-Augmented Generation (PAG)~\citep{richardson2023integrating}, which enhances RAG with user history summaries; Supervised Fine-Tuning (SFT) on the full dataset; and SFT-Ind, which is fine-tuned only on individual task data. Reasoning-enhanced variants of these methods, denoted with a `-R` suffix, are also included. We further compare against advanced techniques. Personalization-focused methods include PPlug~\citep{liu2025llms+}, a plug-and-play approach using specialized user embeddings, and HYDRA-Adapter~\citep{zhuang2024hydra}, for which we use only its adapter version to maintain a consistent retrieval method for fairness. Reasoning-focused baselines include R2P~\citep{luo2025reasoning}, which employs hierarchical reasoning templates, and PrLM~\citep{zhang2025prlm}, which uses a contrastive reward model with reinforcement learning. To situate our method's performance against the frontier of language models, we also include several leading state-of-the-art  LLMs: GPT-4o~\citep{hurst2024gpt}, Gemini-2.5-Pro~\citep{comanici2025gemini}, Qwen3-235B-A22B~\citep{yang2025qwen3}, and DeepSeek-R1~\citep{guo2025deepseek}.

We deliberately exclude methods centered on optimizing the retrieval module, as improving retrieval is not the focus of our research. Additionally, we do not compare against OPPU~\citep{tan2024democratizing}, as its approach requires fine-tuning a unique module for every user and presupposes the availability of extensive user-specific profiles, rendering it infeasible to implement across the full validation set.

\section{Additional Further Analysis}
\label{app:add_further_analysis}

\subsection{Model Training Cost}
The additional training cost relative to SFT-R arises primarily from multi-stage RL optimization. As shown in Table~\ref{tab:training_cost}, TagPR requires 169.7 A100 GPU-hours, compared with 177.7 A100 GPU-hours for the untagged RL baseline and 8.0 A100 GPU-hours for SFT-R. Thus, TagPR requires more computation than SFT alone but remains comparable in cost to an otherwise equivalent RL-based method.

\begin{table*}[htbp]
  \centering
  \resizebox{\textwidth}{!}{
  \begin{tabular}{lccccc}
    \toprule
    \textbf{Method} & \textbf{SFT} & \textbf{PRMU Training} & \textbf{RL} & \textbf{Total Wall-Clock Time} & \textbf{Total A100 GPU-Hours} \\
    \midrule
    SFT-R & 1 h $\times$ 8 A100 & -- & -- & 1 h & 8.0 \\
    TagPR w/o Tagging & 1 h $\times$ 8 A100 & 1 h 40 min $\times$ 1 A100 & 21 h $\times$ 8 A100 & 23 h 40 min & 177.7 \\
    TagPR & 1 h $\times$ 8 A100 & 1 h 40 min $\times$ 1 A100 & 20 h $\times$ 8 A100 & 22 h 40 min & 169.7 \\
    \bottomrule
  \end{tabular}
  }
  \caption{Training cost comparison. Total wall-clock time assumes sequential execution of SFT, PRMU training, and RL.}
  \label{tab:training_cost}
\end{table*}

\subsection{PRMU Reward Model Evaluation}
\label{app:prmu_evaluation}

We evaluate the standalone preference prediction accuracy of PRMU variants on a held-out set. As shown in Table~\ref{tab:prmu_accuracy}, the full PRMU with user embeddings achieves the best accuracy (80.45\%), substantially outperforming an off-the-shelf reward model (69.25\%), confirming the value of both domain-specific fine-tuning and user-aware modeling.

\begin{table}[htbp]
  \centering
  \resizebox{0.85\linewidth}{!}{%
  \begin{tabular}{lc}
    \toprule
    \textbf{Method} & \textbf{Accuracy} \\
    \midrule
    Off-the-shelf Base RM & 69.25\% \\
    Base RM fine-tuned on our dataset & 75.40\% \\
    \rowcolor{aliceblue!60} \textbf{PRMU (full)} & \textbf{80.45\%} \\
    \bottomrule
  \end{tabular}%
  }
  \caption{Preference prediction accuracy of PRMU variants on a held-out set.}
  \label{tab:prmu_accuracy}
\end{table}

\subsection{Hyperparameter Sensitivity Analysis of Reward Function}
\label{app:hyperparameter}

We conducted a sensitivity analysis on the reward function hyperparameters by varying the weights of $\alpha$, $\beta$, and $\gamma$ while keeping all other experimental conditions constant. The results, presented in Table~\ref{tab:hyperparameter_heldout}, indicate that the model remains robust within a reasonable range of parameter variations, with our current configuration yielding the most balanced performance.

\begin{table*}[htbp]
  \centering
      \resizebox{\textwidth}{!}{%
    \begin{tabular}{lcccccccccccc}
    \toprule
    \textbf{Dataset $\rightarrow$} & \multicolumn{2}{c}{\textbf{LaMP-1}} & \multicolumn{2}{c}{\textbf{LaMP-2}} & \multicolumn{2}{c}{\textbf{LaMP-3}} & \multicolumn{2}{c}{\textbf{LaMP-4}} & \multicolumn{2}{c}{\textbf{LaMP-5}} & \multicolumn{2}{c}{\textbf{LaMP-7}} \\
    \cmidrule(lr){1-1}
    \cmidrule(lr){2-3} \cmidrule(lr){4-5} \cmidrule(lr){6-7} \cmidrule(lr){8-9} \cmidrule(lr){10-11} \cmidrule(lr){12-13}
    \textbf{Method} & \textbf{ACC $\uparrow$} & \textbf{F1 $\uparrow$} & \textbf{ACC $\uparrow$} & \textbf{F1 $\uparrow$} & \textbf{MAE $\downarrow$} & \textbf{RMSE $\downarrow$} & \textbf{R-1 $\uparrow$} & \textbf{R-L $\uparrow$} & \textbf{R-1 $\uparrow$} & \textbf{R-L $\uparrow$} & \textbf{R-1 $\uparrow$} & \textbf{R-L $\uparrow$} \\
    \midrule
    \textbf{$\alpha = 1.0, \beta = 1.0, \gamma = 1.0$} & 0.798 & 0.796 & 0.581 & 0.553 & 0.226 & 0.271 & 0.227 & 0.209 & 0.540 & 0.469 & 0.564 & 0.505 \\
    \textbf{$\alpha = 0.2, \beta = 0.2, \gamma = 0.8$} & 0.791 & 0.788 & 0.573 & 0.533 & 0.229 & 0.272 & 0.226 & 0.208 & 0.533 & 0.459 & 0.561 & 0.501 \\
    \textbf{$\alpha = 0.8, \beta = 0.2, \gamma = 0.2$} & 0.792 & 0.790 & 0.576 & 0.538 & 0.228 & 0.272 & 0.228 & 0.210 & 0.540 & 0.467 & 0.565 & 0.506 \\
    \textbf{$\alpha = 0.2, \beta = 0.8, \gamma = 0.2$} & 0.787 & 0.786 & 0.574 & 0.536 & 0.234 & 0.274 & 0.225 & 0.207 & 0.535 & 0.460 & 0.560 & 0.499 \\
    \textbf{$\alpha = 0.8, \beta = 0.8, \gamma = 0.4$} & 0.803 & 0.801 & 0.598 & \textbf{0.559} & \textbf{0.215} & \textbf{0.262} & 0.232 & \textbf{0.215} & \textbf{0.543} & 0.471 & 0.565 & \textbf{0.507} \\
    \rowcolor{aliceblue!60} $\alpha = 0.8, \beta = 0.8, \gamma = 0.2$ (Ours) & \textbf{0.805} & \textbf{0.804} & \textbf{0.599} & 0.558 & 0.218 & 0.263 & \textbf{0.234} & 0.214 & 0.542 & \textbf{0.472} & \textbf{0.566} & 0.506 \\
    \bottomrule
    \end{tabular}%
    }
    \caption{Hyperparameter sensitivity analysis on a held-out validation set (500 instances per task, sampled from unused training data). Our default configuration achieves optimal or near-optimal performance. Best results in \textbf{bold}.}
    \label{tab:hyperparameter_heldout}%
\end{table*}%

\subsection{Semantic Consistency of Generated Tags}
\label{app:semantic_consistency}

To verify that generated tags reflect genuine semantic content rather than memorized syntactic templates, we randomly sampled 500 tag-content pairs from the model's responses on the final test set. We employed Gemini-3-Flash-Preview as an independent judge with the following prompt:

\begin{tcolorbox}[colframe=grey!20!white, colback=grey!3!white, coltitle=black, fonttitle=\bfseries, title=Prompt for Semantic Consistency Evaluation, breakable]
Judge whether the tag name and the content below are semantically consistent, meaning whether the content matches what the tag name expresses. Reply with exactly one word, consistent or inconsistent.\\
\\
Tag name: \{tag\_name\}\\
Content: \{content\}\\
Answer:
\end{tcolorbox}

The evaluation yielded a 100\% consistency rate. This strong semantic alignment indicates that the tags effectively externalize the model's intermediate cognitive steps in a human-interpretable manner, rather than serving as empty structural placeholders.

\begin{figure*}[htbp]
 \centering
\includegraphics[width=0.9\linewidth]{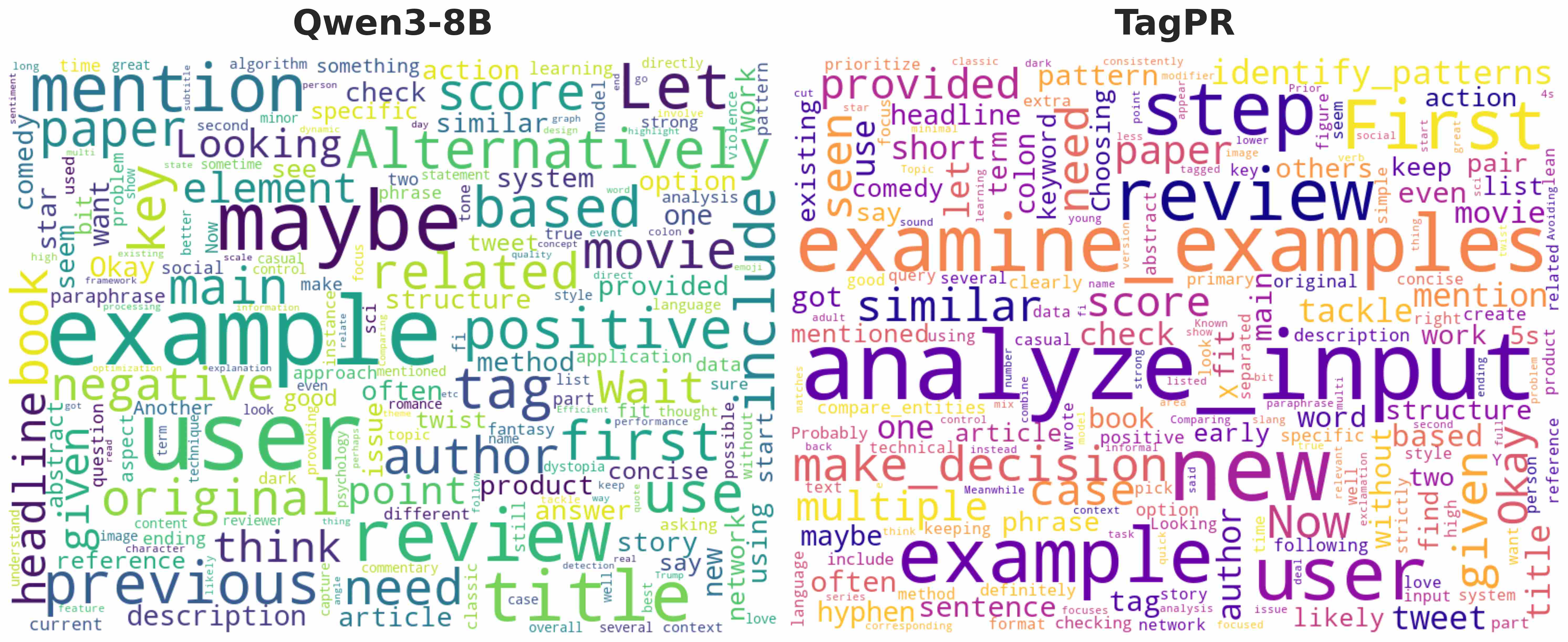}
\caption{Word cloud comparison of reasoning chains from the baseline Qwen3-8B (left) and our TagPR model (right) on the LaMP validation set. TagPR's reasoning is dominated by action-oriented keywords derived from our functional tags.}
\label{fig:wordcloud}
\end{figure*}

\subsection{Case Study}
We present qualitative case studies to illustrate the advanced personalization reasoning of \textbf{TagPR}. The primary case is shown in Figure~\ref{fig:reasoning_comparison}. Additional cases are presented below.

\begin{tcolorbox}[colframe=grey!20!white, colback=grey!3!white, coltitle=black, fonttitle=\bfseries, title=Response from Qwen3-8B for Case 1, breakable]

\textless think\textgreater \\
Okay, let's see. The user provided a list of papers by an author and is asking which of the two given references is related. The options are [1] ``Compressed Linear Algebra for Large-Scale Machine Learning'' and [2] ``Database and information-retrieval methods for knowledge discovery''.

First, I need to look at the titles of the author's papers to see which topic they're focusing on. The author's papers include things like Bayesian specification learning for API errors, distributed Bayesian machine learning, complex object implementations for big data, distributed algorithms for covariance matrices, grading in MOOCs, learning to grade student programs, comparing machine learning platforms, simulating database-valued Markov chains, and a model-agnostic framework for spatial anomaly detection.

Looking at these, several themes emerge: ...

Now, the two options are about compressed linear algebra for machine learning and database/information retrieval methods for knowledge discovery. ... Wait, the user's question is: ``which reference is related?'' given the list of the author's papers and the two options. So the user is asking which of the two options is related to the author's work.

So, the author's papers include topics like distributed Bayesian machine learning, big data systems, database simulations (SimSQL), and machine learning platforms. The first option, [1], is about compressed linear algebra for large-scale ML, which is a machine learning technique. The second option, [2], is about database and information retrieval methods for knowledge discovery. ... But the user is asking which of the two references is related. The answer is likely [2], but I'm not 100\% certain. Alternatively, maybe [1] is related because the author's work on large-scale machine learning (like the paper on distributed algorithms) could be connected to compressed linear algebra. But the user's options are two references, and the answer is to choose one. Given that the author has a paper on SimSQL (database-related), [2] is about databases and information retrieval, so [2] is more directly related. Therefore, the answer is [2].\\
\textless /think\textgreater

[2]

\end{tcolorbox}

\begin{tcolorbox}[colframe=grey!20!white, colback=grey!3!white, coltitle=black, fonttitle=\bfseries, title=Response from TagPR for Case 1, breakable]

\textless{}think\textgreater{}\\
\textless{}analyze\_input\textgreater{}\\
Okay, let me try to figure out which reference is related to the author's papers. The user provided a list of papers the author has written and two options for references: [1] ``Compressed Linear Algebra for Large-Scale Machine Learning'' and [2] ``Database and information-retrieval methods for knowledge discovery''.\\
\textless{}/analyze\_input\textgreater{}\\
\textless{}examine\_examples\textgreater{}\\
First, I need to look at the titles of the author's papers to see what themes they're working on. Let me list them out again:\\
1. ``Bayesian specification learning for finding API usage errors''

...

9. ``A Model-Agnostic Framework for Fast Spatial Anomaly Detection''\\
\textless{}/examine\_examples\textgreater{}\\
\textless{}identify\_patterns\textgreater{}\\
Looking at these, several themes emerge. There's a strong focus on machine learning, especially distributed and large-scale aspects. Papers like ``The BUDS Language for Distributed Bayesian Machine Learning'' and ``A comparison of platforms for implementing and running very large scale machine learning algorithms'' clearly relate to machine learning systems. Also, ``Distributed Algorithms for Computing Very Large Thresholded Covariance Matrices'' and ``Compressed Linear Algebra for Large-Scale Machine Learning'' (which is reference [1]) might be connected because they both deal with large-scale computations in machine learning.\\
\textless{}/identify\_patterns\textgreater{}\\
\textless{}evaluate\_reference\textgreater{}\\
Then there's the database-related paper: ...\\
\textless{}/evaluate\_reference\textgreater{}\\
\textless{}compare\_entities\textgreater{}\\
Now, comparing the two references: [1] is about compressed linear algebra for machine learning, which aligns with papers on large-scale ML algorithms and distributed computing. [2] is about database and information-retrieval methods for knowledge discovery, which connects to the database-valued Markov chains paper.\\
\textless{}/compare\_entities\textgreater{}\\
\textless{}synthesize\_findings\textgreater{}\\
But the author has multiple papers on machine learning, especially distributed and large-scale. ...\\
\textless{}/synthesize\_findings\textgreater{}\\
\textless{}make\_decision\textgreater{}\\
So, reference [1] seems more broadly related to the author's work in machine learning systems and large-scale algorithms. Reference [2] is only tangentially related through one paper, while [1] connects to multiple papers.\\
\textless{}/make\_decision\textgreater{}\\
\textless{}/think\textgreater{}\\
{[}1{]}

\end{tcolorbox}

\begin{tcolorbox}[colframe=grey!20!white, colback=grey!3!white, coltitle=black, fonttitle=\bfseries, title=Response from Qwen3-8B for Case 2, breakable]

\textless think\textgreater \\
Okay, let's see. The user wants me to paraphrase the tweet ``@Ylsoccermomof3, I appreciate it and I'll enjoy my double cheeseburger.'' without any explanation before or after. \\
First, I need to understand the original message. The person is thanking @Ylsoccermomof3 and mentioning they'll enjoy a double cheeseburger. The key elements are gratitude and the mention of the food. \\
Now, paraphrasing means rewording it while keeping the same meaning. Let me think of synonyms. ``Appreciate'' could be ``thank'' or ``grateful for.'' ``Enjoy'' might be ``savor'' or ``look forward to.'' ``Double cheeseburger'' is specific, so maybe keep it as is or use ``big cheeseburger'' for a slight variation. \\
Also, the structure: the original starts with the handle, then the message. The user's example response uses ``Thanks, @Ylsoccermomof3, and I'll savor my double cheeseburger.'' So they changed ``appreciate'' to ``Thanks'' and ``enjoy'' to ``savor.'' That's a good approach. \\
I should check if there are other ways to phrase it. Maybe ``I'm grateful, @Ylsoccermomof3, and I'll relish my double cheeseburger.'' But ``relish'' might be a bit more formal. The original is casual, so ``savor'' fits better. \\
Another option: ``Thanks, @Ylsoccermomof3, and I'll enjoy my big cheeseburger.'' But ``big'' might not be the same as ``double.'' The user's example used ``double,'' so maybe stick with that. \\
I think the user's example is good, but maybe check for any other possible variations. However, the user wants a paraphrase, not necessarily the same as the example. Let me make sure the structure is correct. The handle is included, then the message. \\
...
\\
Thanks, @Ylsoccermomof3, and I'll savor my double cheeseburger.

\end{tcolorbox}

\begin{tcolorbox}[colframe=grey!20!white, colback=grey!3!white, coltitle=black, fonttitle=\bfseries, title=Response from TagPR for Case 2, breakable]

\textless think\textgreater\\
\textless analyze\_input\textgreater\\
Okay, the user has given me a task to paraphrase a specific tweet based on patterns observed from several example tweets they provided. Let me break this down carefully.\\
\textless /analyze\_input\textgreater\\
\textless examine\_examples\textgreater\\
First, I need to analyze the patterns in the example tweets they shared. Looking at these:\\
- They all start with "@username" followed by a casual greeting or statement\\
- The language is very conversational with contractions ("you're", "it's")\\
- There's a mix of personal updates ("my cat is having kittens") and light-hearted interactions\\
- Some include multiple @mentions\\
- The tone is friendly and informal\\
\textless /examine\_examples\textgreater\\
\textless identify\_patterns\textgreater\\
The key pattern I notice is the casual, conversational style with:\\
- Use of "I" statements\\
- Simple sentence structures\\
- No formal punctuation beyond basic commas\\
- Often includes personal context ("I just woke up", "I think everyone has")\\
\textless /identify\_patterns\textgreater\\
\textless analyze\_input\textgreater\\
Now for the tweet to paraphrase: "@Ylsoccermomof3, I appreciate it and I'll enjoy my double cheeseburger."\\
\textless /analyze\_input\textgreater\\
\textless identify\_patterns\textgreater\\
Hmm... the original has:\\
- Formal "I'll" instead of "I'm gonna"\\
- "double cheeseburger" which is slightly more specific than typical casual phrasing\\
- No contractions in "I'll" (though it's common in speech)\\
\textless /identify\_patterns\textgreater\\

...\\

\textless verify\_conclusion\textgreater\\
So "I'm gonna" is better than "I'll" for that casual vibe. And "enjoy" could become "devour" like in the @Jintanut example ("very hungry") but "enjoy" is already good.\\
\textless /verify\_conclusion\textgreater\\
\textless formulate\_conclusion\textgreater\\
Final decision: Keep the structure but make it sound like a real person talking. Add "thanks" to match the gratitude pattern, use "gonna" for casualness, and keep the @mention.\\
\textless /formulate\_conclusion\textgreater\\
\textless /think\textgreater\\
@Ylsoccermomof3, thanks for the encouragement and I'm gonna enjoy my double cheeseburger.

\end{tcolorbox}

\section{Supplement for Generalization Assessment}
\label{app:generalization_assessment}

\subsection{Generalization Assessment on LongLaMP}

    We further evaluated the generalization capabilities of our model using the LongLaMP dataset. Characterized by extensive user history profiles and a requirement for long-form text generation, LongLaMP poses a significant challenge to personalized reasoning. We focused on three tasks: TopicWriting, ProductReview, and AbstractReviewTitle. Due to computational constraints, we randomly sampled 10\% of the test set for each task and reported performance using ROUGE-1 (R-1) and ROUGE-L (R-L).

We assessed zero-shot generalization by directly applying the model trained on the LaMP dataset to the LongLaMP dataset. As shown in Table~\ref{tab:tab_generalization_long_pgarph}, our model demonstrated robust zero-shot generalization capabilities.

\subsection{Generalization Assessment on PGraphRAG}

We further evaluated the model's generalization capabilities using the PGraphRAG dataset. This dataset is characterized by extreme sparsity, where the majority of user profiles contain either zero or only one, posing a significant challenge for personalized reasoning. We focused specifically on the AmazonReviewTitle subtask. Since historical profiles are the foundation of personalization, we restricted our evaluation to test users with a profile history length of at least one, excluding neighbor profiles information. By directly applying our trained model, we demonstrated its robust personalized reasoning capabilities even in data-sparse scenarios, as shown in Table~\ref{tab:tab_generalization_long_pgarph}.

\begin{table*}[htbp]
  \centering

    \resizebox{0.9\textwidth}{!}{%
    \begin{tabular}{lcccccccc}
    \toprule
    \textbf{Dataset $\rightarrow$} & \multicolumn{2}{c}{\textbf{TopicWriting}} & \multicolumn{2}{c}{\textbf{ProductReview}} & \multicolumn{2}{c}{\textbf{AbstractGeneration}} & \multicolumn{2}{c}{\textbf{AmazonReviewTitle}} \\
    \cmidrule(lr){1-1}
    \cmidrule(lr){2-3} \cmidrule(lr){4-5} \cmidrule(lr){6-7} \cmidrule(lr){8-9}
    \textbf{Method} & \textbf{R-1 $\uparrow$} & \textbf{R-L $\uparrow$} & \textbf{R-1 $\uparrow$} & \textbf{R-L $\uparrow$} & \textbf{R-1 $\uparrow$} & \textbf{R-L $\uparrow$} & \textbf{R-1 $\uparrow$} & \textbf{R-L $\uparrow$} \\
    \midrule
    \textbf{Qwen3-8B-Instruct} & 0.282  & 0.134  & 0.342  & 0.154  & 0.382  & 0.203  & 0.172  & 0.163  \\
    \textbf{Qwen3-8B-Thinking (Base)} & 0.271  & 0.124  & 0.321  & 0.150  & 0.351  & 0.182  & 0.179  & 0.165  \\
    \textbf{Qwen3-32B-Instruct} & 0.292  & 0.134  & 0.354  & 0.159  & 0.384  & 0.202  & 0.171  & 0.161  \\
    \textbf{Qwen3-32B-Thinking} & 0.275  & 0.119  & 0.320  & 0.144  & 0.346  & 0.177  & 0.198  & 0.191  \\
    \textbf{GPT-4o} & 0.294  & 0.140  & 0.330  & 0.157  & 0.372  & 0.200  & 0.140  & 0.136  \\
    \textbf{Gemini-2.5-Pro} & 0.270  & 0.132  & 0.372  & 0.171  & 0.398  & 0.215  & 0.204  & 0.196  \\
    \rowcolor{aliceblue!60} \textbf{TagPR (Our)} & \textbf{0.303} & \textbf{0.154} & \textbf{0.377} & \textbf{0.175} & \textbf{0.404} & \textbf{0.221} & \textbf{0.210} & \textbf{0.202} \\
    \bottomrule
    \end{tabular}%
    }
    \caption{Zero-shot generalization performance on partial test sets of LongLaMP and PGraphRAG. 
  We report ROUGE-1 (R-1) and ROUGE-L (R-L) scores. 
  The best results are in \textbf{bold}. Our TagPR demonstrates superior performance.}
  \label{tab:tab_generalization_long_pgarph}%
\end{table*}%

\section{Supplement for Robustness Assessment}
\label{app:robustness}

\begin{figure*}
 \centering
\includegraphics[width=1.0\linewidth]{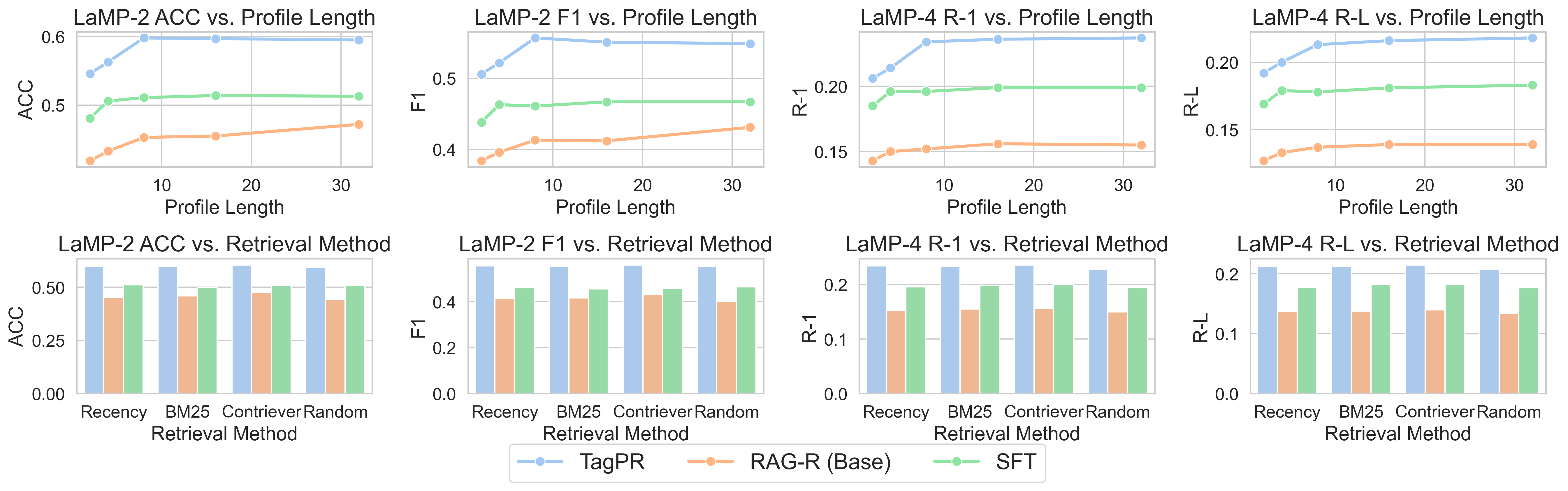}
\caption{ Robustness assessment of TagPR on LaMP-2 and LaMP-4. \textbf{Top:} Performance across varying profile lengths. \textbf{Bottom:} Performance across different retrieval methods. TagPR consistently outperforms baselines, demonstrating high data efficiency and resilience to retrieval quality.}
\label{fig:roubustness_assessment}
\end{figure*}

\begin{figure*}
 \centering
\includegraphics[width=1.0\linewidth]{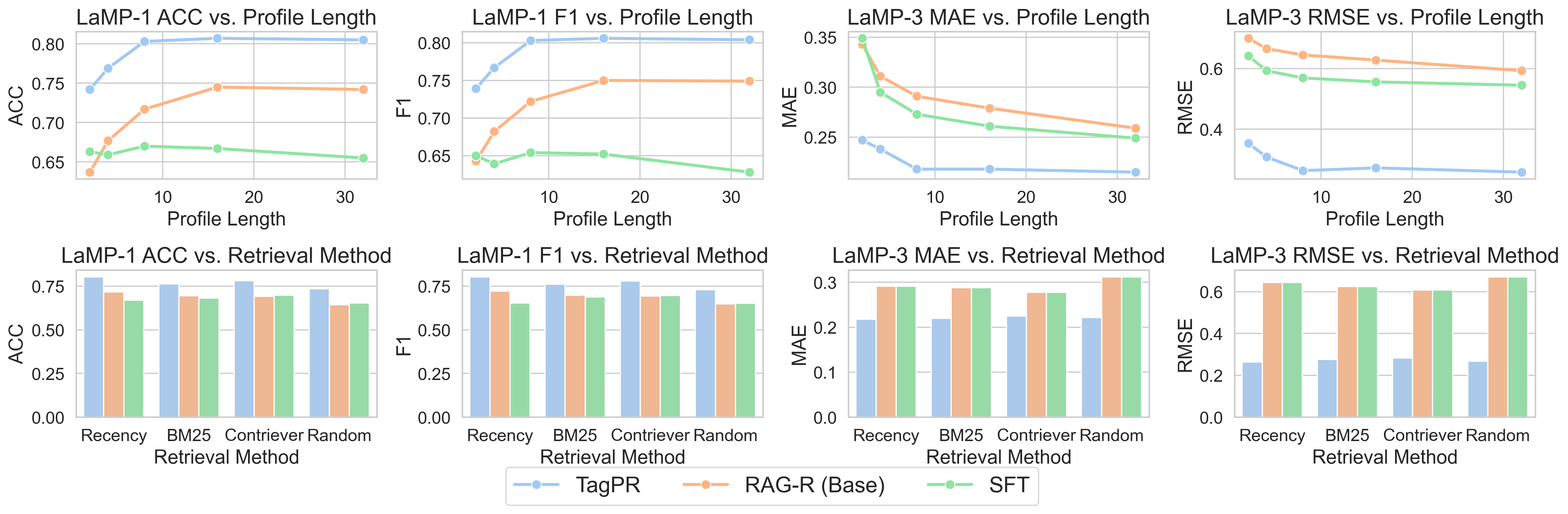}
\caption{ Robustness assessment of TagPR on LaMP-1 and LaMP-3. \textbf{Top:} Performance across varying profile lengths. \textbf{Bottom:} Performance across different retrieval methods.}
\label{fig:robustness_assessment_set2_v2}
\end{figure*}

\begin{figure*}
 \centering
\includegraphics[width=1.0\linewidth]{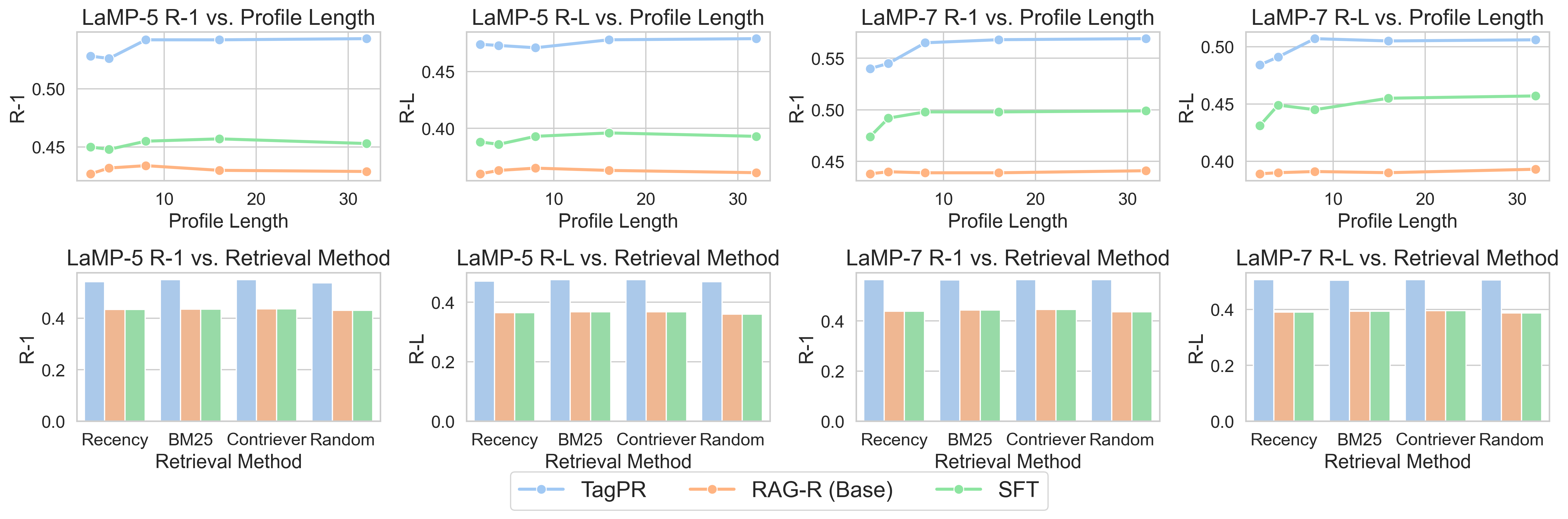}
\caption{ Robustness assessment of TagPR on LaMP-5 and LaMP-7. \textbf{Top:} Performance across varying profile lengths. \textbf{Bottom:} Performance across different retrieval methods.}
\label{fig:robustness_assessment_set3_v2}
\end{figure*}

This section provides supplementary results for the robustness assessment discussed in the main paper. Figure~\ref{fig:robustness_assessment_set2_v2} and Figure~\ref{fig:robustness_assessment_set3_v2} illustrate the performance of \textbf{TagPR} against the SFT and Base baselines on the LaMP-1, LaMP-3, LaMP-5, and LaMP-7 tasks, complementing the results for LaMP-2 and LaMP-4 shown in Figure~\ref{fig:roubustness_assessment}.

As demonstrated in the figure, the conclusions from the main text hold true across these additional datasets. \textbf{TagPR} consistently achieves superior performance, showcasing high data efficiency by reaching a strong performance level with only a few user interactions. Furthermore, its advantage is maintained across all profile retrieval methods, including random selection, which underscores the robustness of our framework.

\section{Supplement for Tagged Reasoning Chains Construction}

\subsection{Prompts Details}

\subsubsection{The Prompt for LLM Filter}
\label{app:prompt_llm_filter}

\begin{tcolorbox}[colframe=grey!20!white, colback=grey!3!white, coltitle=black, fonttitle=\bfseries, title=Prompt for LLM Filter, breakable]
\textbf{\# Role} \par
You are an AI expert specializing in evaluating the Chain-of-Thought (CoT) quality of large language models. Your task is to provide a comprehensive and objective evaluation of the model's Chain-of-Thought quality based on the provided question and the model's response.
\vspace{\baselineskip}

\textbf{\# Task Description} \par
I will provide you with a ``Question" and a ``Model Response" generated by a large language model for that question. The ``Model Response" contains detailed reasoning steps and the final answer. Please evaluate the quality of the Chain-of-Thought in this ``Model Response" according to the following evaluation dimensions, and strictly output the result in the specified JSON format.
\vspace{\baselineskip}

\textbf{\# Evaluation Dimensions}
\begin{enumerate}
    \item \textbf{Logical Coherence:} Is there a clear logical connection between the reasoning steps? Are there any logical leaps or contradictions? (1-5 points)
    \item \textbf{Step Accuracy:} Is every step in the reasoning chain accurate? Are there any factual errors or calculation mistakes? (1-5 points)
    \item \textbf{Reasoning Completeness:} Does the Chain-of-Thought cover all the key steps required to solve the problem? Are there any omissions? (1-5 points)
    \item \textbf{Relevance to the Question:} Is the entire thought process closely centered around the original question? Is there any redundant or off-topic reasoning? (1-5 points)
\end{enumerate}
\vspace{\baselineskip}

\textbf{\# Input Data} \par
\textbf{[Question]} \par
\{question\}
\vspace{0.5\baselineskip}

\textbf{[Model Response]} \par
\{response\}
\vspace{\baselineskip}

\textbf{\# Output Format} \par
Please strictly follow the JSON format below for your evaluation results. Ensure the output is a complete and syntactically correct JSON object. Do not add any additional explanations or text before or after the JSON code block.

\begin{lstlisting}[style=jsonstyle]
{
  "evaluation_report": {
  "detailed_assessment": [
      {    
        "dimension": "Logical Coherence",
        "reasoning": "[Provide an explanation of the pros and cons for this dimension]",
        "score": "[Enter an integer score from 1-5 here]"
      },
      {
        "dimension": "Step Accuracy",
        "reasoning": "[Provide an explanation of the pros and cons for this dimension, and explicitly point out any errors if they exist]",
        "score": "[Enter an integer score from 1-5 here]"
      },
      {
        "dimension": "Reasoning Completeness",
        "reasoning": "[Provide an explanation of the pros and cons for this dimension, and explicitly point out any omissions if they exist]",
        "score": "[Enter an integer score from 1-5 here]"
      },    
      {
        "dimension": "Relevance to the Question",
        "reasoning": "[Provide an explanation of the pros and cons for this dimension, such as the presence of redundant information]",
        "score": "[Enter an integer score from 1-5 here]"
      }
    ],
  "summary": {
      "total_score": "[Enter the total score, between 1 and 20, which is the sum of the scores from each dimension]"
    }
  }
}
\end{lstlisting}
\end{tcolorbox}

\subsubsection{The Prompt for Exploratory Tagging}
\label{app:prompt_exploratory_tagging}

\begin{tcolorbox}[colframe=grey!20!white, colback=grey!3!white, coltitle=black, fonttitle=\bfseries, title=Prompt for Exploratory Tagging, breakable]
\textbf{Role:} You are an expert specializing in understanding and analyzing the thought processes of AI. Your task is to carefully review a given question and the ``Chain-of-Thought" generated by an AI model to answer it. You will then break down this Chain-of-Thought into meaningful segments and assign an XML-style tag to each segment that best describes its function.
\vspace{\baselineskip}

\textbf{Task:}

Based on the user-provided \textbf{[Question]} and the model-generated \textbf{[Chain-of-Thought]}, please complete the following steps:
\begin{enumerate}
    \item \textbf{Analyze the Question and Chain-of-Thought:} Deeply understand the core requirements of the question and how the Chain-of-Thought progressively derives the final answer.
    \item \textbf{Segment the Chain-of-Thought:} Break down the entire Chain-of-Thought into multiple logically coherent steps or stages. Each step should represent a distinct function in the thought process.
    \item \textbf{Tag Each Segment:} For each segmented piece, create an XML-style tag (e.g., \texttt{<tag\_name>...</tag\_name>}). The tag name should accurately summarize the role this segment plays in the overall thought process.
    \item \textbf{Output the Result:} Output only the tagged Chain-of-Thought text.
\end{enumerate}
\vspace{\baselineskip}

\textbf{Guidelines and Tag Suggestions:}
\begin{itemize}
    \item \textbf{Tags should reflect function:} The tag name should describe ``what this segment is doing", not simply restate its content.
    \item \textbf{Maintain consistency:} Try to use the same tags for similar functions.
    \item \textbf{Tags should be concise and clear:} Use short, easy-to-understand words for tag names (English is recommended for easier programmatic processing).
\end{itemize}
\vspace{\baselineskip}

\textbf{Example:}

\textbf{[Question]:} \par
History: ``When brilliant video game maker Flynn hacks the mainframe of his ex-employer\dots" $\rightarrow$ Tag: ``classic" and, ``When an insatiable great white shark terrorizes the townspeople\dots" $\rightarrow$ Tag: ``classic" and, ``Jake Blues, just released from prison, puts his old band back together\dots" $\rightarrow$ Tag: ``classic" and, \dots(other examples of ``classic" and ``true story")\dots \par
Description: ``In the years before the Second World War, a tomboyish postulant at an Austrian abbey is hired as a governess in the home of a widowed naval captain with seven children, and brings a new love of life and music into the home." \par
Candidate Tags: [sci-fi, based on a book, comedy, action, \dots , classic, \dots , true story]
\vspace{0.5\baselineskip}

\textbf{[Chain-of-Thought to be tagged]:} \par
First, I analyzed the historical data provided by the user. This data shows descriptions of several movies and their corresponding tags. \par
I found that the movies tagged as ``classic" (like \textit{Tron}, \textit{Jaws}, \textit{The Blues Brothers}, etc.) are all well-known and beloved films released in the 20th century. Movies tagged as ``true story" (like \textit{Apollo 13}) are based on real events. This reveals a pattern: the ``classic" tag points to older films with a significant era-defining and broad cultural impact. \par
Next, I analyzed the new movie description to be tagged: ``In the years before the Second World War...". This description points to the famous movie \textit{The Sound of Music}. \par
\textit{The Sound of Music}, released in 1965, tells a household story and achieved immense success and cultural influence worldwide, perfectly matching the characteristics of a ``classic" movie that I summarized from the historical data. It is not a sci-fi film, and while it might be partially based on real experiences, its core identity is its classic status. \par
Therefore, among all candidate tags, ``classic" is the most appropriate.
\vspace{0.5\baselineskip}

\textbf{[Your Output]:}
\begin{lstlisting}[style=xmlstyle]
<analyze_user_history>
First, I analyzed the historical data provided by the user. This data shows descriptions of several movies and their corresponding tags.
</analyze_user_history>
<identify_pattern>
I found that the movies tagged as "classic" (like *Tron*, *Jaws*, *The Blues Brothers*, etc.) are all well-known and beloved films released in the 20th century. Movies tagged as "true story" (like *Apollo 13*) are based on real events. This reveals a pattern: the "classic" tag points to older films with a significant era-defining and broad cultural impact.
</identify_pattern>
<analyze_current_query>
Next, I analyzed the new movie description to be tagged: "In the years before the Second World War...". This description points to the famous movie *The Sound of Music*.
</analyze_current_query>
<compare_query_with_history>
*The Sound of Music*, released in 1965, tells a household story and achieved immense success and cultural influence worldwide, perfectly matching the characteristics of a "classic" movie that I summarized from the historical data. It is not a sci-fi film, and while it might be partially based on real experiences, its core identity is its classic status.
</compare_query_with_history>
<final_conclusion>
Therefore, among all candidate tags, "classic" is the most appropriate.
</final_conclusion>
\end{lstlisting}
\vspace{\baselineskip}

\textbf{Now, according to the rules above, please add tags to the [Question] and [Chain-of-Thought] provided below:}
\vspace{0.5\baselineskip}

\textbf{[Question]:} \par
\{question\}
\vspace{0.5\baselineskip}

\textbf{[Chain-of-Thought to be tagged]:} \par
\{chain\_of\_thought\}

\end{tcolorbox}

\subsubsection{The Prompt for Restricted Tagging}
\label{app:prompt_restricted_tagging}

\begin{tcolorbox}[colframe=grey!20!white, colback=grey!3!white, coltitle=black, fonttitle=\bfseries, title=Prompt for Restricted Tagging, breakable]
\textbf{Role:} You are an expert specializing in understanding and analyzing the thought processes of AI. Your task is to carefully review a given question and the ``Chain-of-Thought" generated by an AI model to answer it. You will then break down this Chain-of-Thought into meaningful segments and assign an XML-style tag to each segment that best describes its function.
\vspace{\baselineskip}

\textbf{Task:}

Based on the user-provided \textbf{[Question]} and the model-generated \textbf{[Chain-of-Thought]}, please complete the following steps:
\begin{enumerate}
    \item \textbf{Analyze the Question and Chain-of-Thought:} Deeply understand the core requirements of the question and how the Chain-of-Thought progressively derives the final answer.
    \item \textbf{Segment the Chain-of-Thought:} Break down the entire Chain-of-Thought into multiple logically coherent steps or stages. Each step should represent a distinct function in the thought process.
    \item \textbf{Tag Each Segment:} For each segmented piece, create an XML-style tag (e.g., \texttt{<tag\_name>...</tag\_name>}). The tag name must be chosen exclusively from the mandatory list provided below.
    \item \textbf{Output the Result:} Output only the tagged Chain-of-Thought text.
\end{enumerate}
\vspace{\baselineskip}

\textbf{Mandatory Tag Set and Definitions:}

You \textbf{must} use \textbf{only} the tags from the following list. Choose the tag that best describes the function of each segment.
\begin{itemize}
    \item \texttt{\textbf{analyze\_input}}: Analyzes the initial user question or task description to understand the goal.
    \item \texttt{\textbf{examine\_examples}}: Examines specific, individual pieces of evidence, data points, or examples provided.
    \item \texttt{\textbf{identify\_patterns}}: Summarizes findings from one or more examples to find a common rule, pattern, or theme.
    \item \texttt{\textbf{evaluate\_reference}}: Assesses how the input aligns with a specific, external piece of reference material.
    \item \texttt{\textbf{compare\_entities}}: Performs a direct comparison between two or more items to determine their similarities, differences, or which is superior.
    \item \texttt{\textbf{synthesize\_findings}}: Consolidates all prior analysis and comparisons into a comprehensive summary before making a final choice.
    \item \texttt{\textbf{make\_decision}}: Commits to a specific, final choice or action.
    \item \texttt{\textbf{verify\_conclusion}}: Performs a final check on the decision to ensure it is logical, consistent, and accurate.
    \item \texttt{\textbf{formulate\_conclusion}}: Constructs the final, complete answer or statement based on the decision made.
\end{itemize}
\vspace{\baselineskip}

\textbf{Example:}

\textbf{[Question]:} \par
History: ``When brilliant video game maker Flynn hacks the mainframe of his ex-employer\dots" $\rightarrow$ Tag: ``classic" and, ``When an insatiable great white shark terrorizes the townspeople\dots" $\rightarrow$ Tag: ``classic" and, ``Jake Blues, just released from prison, puts his old band back together\dots" $\rightarrow$ Tag: ``classic" and, \dots(other examples of ``classic" and ``true story")\dots \par
Description: ``In the years before the Second World War, a tomboyish postulant at an Austrian abbey is hired as a governess in the home of a widowed naval captain with seven children, and brings a new love of life and music into the home." \par
Candidate Tags: [sci-fi, based on a book, comedy, action, \dots , classic, \dots , true story]
\vspace{0.5\baselineskip}

\textbf{[Chain-of-Thought to be tagged]:} \par
First, I analyzed the historical data provided by the user. This data shows descriptions of several movies and their corresponding tags. \par
I found that the movies tagged as ``classic" (like \textit{Tron}, \textit{Jaws}, \textit{The Blues Brothers}, etc.) are all well-known and beloved films released in the 20th century. Movies tagged as ``true story" (like \textit{Apollo 13}) are based on real events. This reveals a pattern: the ``classic" tag points to older films with a significant era-defining and broad cultural impact. \par
Next, I analyzed the new movie description to be tagged: ``In the years before the Second World War...". This description points to the famous movie \textit{The Sound of Music}. \par
\textit{The Sound of Music}, released in 1965, tells a household story and achieved immense success and cultural influence worldwide, perfectly matching the characteristics of a ``classic" movie that I summarized from the historical data. It is not a sci-fi film, and while it might be partially based on real experiences, its core identity is its classic status. \par
Therefore, among all candidate tags, ``classic" is the most appropriate.
\vspace{0.5\baselineskip}

\textbf{[Your Output]:}
\begin{lstlisting}[style=xmlstyle]
<examine_examples>
First, I analyzed the historical data provided by the user. This data shows descriptions of several movies and their corresponding tags.
</examine_examples>
<identify_patterns>
I found that the movies tagged as "classic" (like *Tron*, *Jaws*, *The Blues Brothers*, etc.) are all well-known and beloved films released in the 20th century. Movies tagged as "true story" (like *Apollo 13*) are based on real events. This reveals a pattern: the "classic" tag points to older films with a significant era-defining and broad cultural impact.
</identify_patterns>
<analyze_input>
Next, I analyzed the new movie description to be tagged: "In the years before the Second World War...". This description points to the famous movie *The Sound of Music*.
</analyze_input>
<compare_entities>
*The Sound of Music*, released in 1965, tells a household story and achieved immense success and cultural influence worldwide, perfectly matching the characteristics of a "classic" movie that I summarized from the historical data. It is not a sci-fi film, and while it might be partially based on real experiences, its core identity is its classic status.
</compare_entities>
<make_decision>
Therefore, among all candidate tags, "classic" is the most appropriate.
</make_decision>
\end{lstlisting}
\vspace{\baselineskip}

\textbf{Now, according to the rules above, please add tags to the [Question] and [Chain-of-Thought] provided below:}
\vspace{0.5\baselineskip}

\textbf{[Question]:} \par
\{question\}
\vspace{0.5\baselineskip}

\textbf{[Chain-of-Thought to be tagged]:} \par
\{chain\_of\_thought\}

\end{tcolorbox}

\subsection{Refined Primary Tags Set}
\label{app:refined_tags}
\begin{figure*}
 \centering
\includegraphics[width=0.73\linewidth]{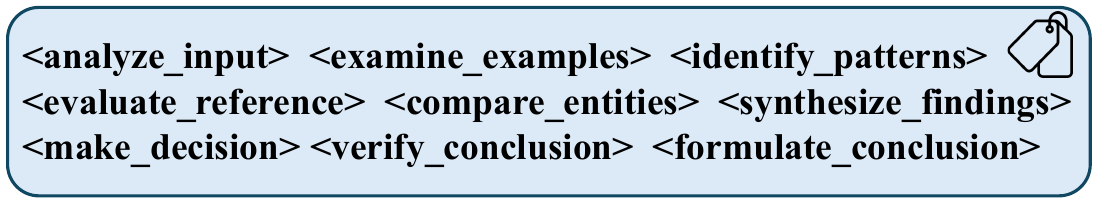}
\caption{The refined set of nine primary tags used for annotating reasoning chains. These tags represent the most salient reasoning patterns identified through our clustering analysis.}
\label{fig:main_tags_set}
\end{figure*}

The final set of primary tags derived from our clustering procedure is listed in Figure~\ref{fig:main_tags_set}. These tags were used to annotate the reasoning chains in our dataset. To determine the number of primary tags ($K$) during the clustering phase, we analyzed the Within-Cluster Sum of Squares (WCSS) and selected $K=9$ to balance cluster compactness with diminishing returns.

\section{New Constructed Personalization Benchmark}
\label{app:dianping_concise}

To evaluate zero-shot, cross-lingual generalization, we built a benchmark from Dianping, a prominent Chinese user-generated content platform. This appendix details its construction.

\subsection{Data and User Profile Creation}
We collected public posts from Dianping and applied rigorous filtering to retain high-quality content, removing short posts, duplicates, and advertisements. From this cleaned dataset, we selected 1,000 users with extensive post histories. 

For each user, a profile representing their personal writing style was constructed from their 8 most recent posts (title and content). The 9th most recent post was held out as the ground truth for our evaluation tasks, ensuring a strict zero-shot setting where the test data is unseen.

\subsection{Task Formulation}
The benchmark consists of three distinct tasks, with one instance per user for each task, totaling 3,000 evaluation instances. All tasks are conditioned on the user's 8-post profile. As in the LaMP dataset, we use the ROUGE-1 and ROUGE-L metrics for evaluation.

\textbf{Dianping-Content (Title \(\rightarrow\) Content):} Given the title of the held-out post, the model must generate the full post content in the user's specific style.

\textbf{Dianping-Title (Content \(\rightarrow\) Title):} The inverse task, where the model generates a stylistically appropriate title from the held-out post's content.

\textbf{Dianping-Paraph (Generic \(\rightarrow\) Stylized Post):} This task measures stylistic transfer. For each user's held-out post, we first used a general-purpose LLM (GPT-4o) to generate a neutral, generic version based on the original content. The model's task is to rewrite this generic text to match the user's unique style, with the user's original post as the target.

\subsection{Benchmark Statistics}
Key statistics of the final benchmark are summarized in Table~\ref{tab:dianping_stats_concise}.

\begin{table}[!ht]
 \center
 \small
  \resizebox{\linewidth}{!}{
  \begin{tabular}{llrrr}
  	\toprule
        {\textbf{Task}} & {\textbf{Task Type}} & \textbf{\#Test} & \textbf{\#Classes} \\
        \midrule
        Dianping-Content & Text generation & 1000 & - \\
        Dianping-Title & Text generation & 1000 & - \\    
        Dianping-Paraph & Text generation & 1000 & - \\
 \bottomrule
  \end{tabular}
  }
   \caption{Data statistics of the new constructed personalization benchmark.} 
  \label{tab:dianping_stats_concise}
\end{table}

\end{document}